\documentclass[runningheads]{llncs}

\usepackage{eccv}

\usepackage{eccvabbrv}

\usepackage{graphicx}
\usepackage{booktabs}

\usepackage[accsupp]{axessibility}  
\usepackage{hyperref}

\usepackage{orcidlink}
\usepackage[numbers]{natbib}
\usepackage{amsmath}
\usepackage{graphicx}
\usepackage{amsfonts}
\usepackage{color}
\usepackage{colortbl}
\usepackage{tabularx}
\usepackage{booktabs}
\usepackage{multirow}
\usepackage{float}
\usepackage{wrapfig}
\usepackage{tcolorbox}
\usepackage{fontawesome5}
\tcbuselibrary{listings, breakable, skins}
\usepackage{enumitem}
\makeatletter
\renewcommand\subsubsection{\@startsection{subsubsection}{3}{\z@}%
  {6\p@}{3\p@}%
  {\normalfont\normalsize\bfseries}}
\makeatother

\tcbset{
  promptbox/.style={
    enhanced, breakable,
    colback=gray!6, colframe=black!70,
    fonttitle=\bfseries\small,
    attach boxed title to top left={yshift=-2mm, xshift=6mm},
    boxed title style={colback=black!75, colframe=black!75, sharp corners},
    coltitle=white,
    sharp corners=south,
    top=4mm, left=3mm, right=3mm, bottom=3mm,
    listing only,
    listing options={
      basicstyle=\ttfamily\scriptsize,
      breaklines=true,
      columns=fullflexible,
      keepspaces=true,
    }
  }
}

\begin{document}

\title{\textbf{RoboTALES}: Learning Reasoning-Guided Robot Policies via Task-Aligned Simulated Futures}
\titlerunning{RoboTALES}


\author{Hanan Gani \and
Tejal Kulkarni \and Madhoolika Chodavarapu
\and Nicklas Hansen \and Manmohan Chandraker}
\institute{University of California, San Diego, CA 92093, USA \\
Correspondence to: \email{hgani@ucsd.edu}}
\authorrunning{Hanan Gani et al.}

\maketitle

\begin{figure}[t] 
    \centering
    \includegraphics[width=1.0\textwidth]{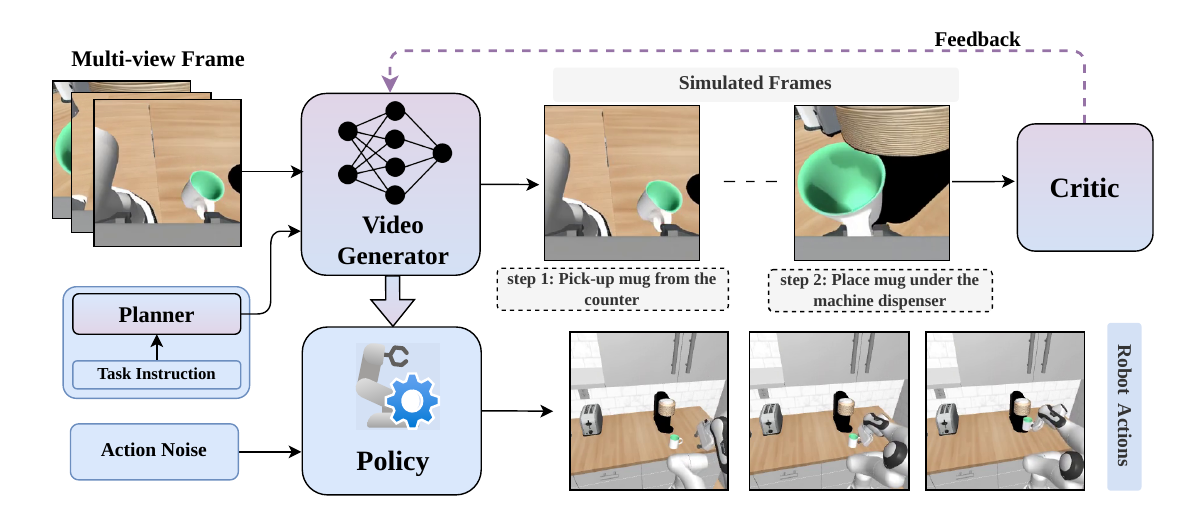}
    \caption{\small \textbf{Overview.} 
An LLM Planner decomposes the task instruction (\emph{e.g.}, ``make a coffee'') 
into ordered subtasks that condition the video generator to generate future frames 
for each milestone. A frozen VLM Critic scores these imagined rollouts against 
the goal instruction, steering the video generator model toward semantically aligned futures. 
The resulting goal-conditioned features drive the policy to produce precise robot 
actions for long-horizon manipulation.}
    \label{fig:teaser}
\vspace{-2em}
\end{figure}

\begin{abstract}
Pretrained video generative models are promising backbones for visuomotor control, but their imagined futures often drift from task intent and are not reliably action-conditional. As a result, these models can be difficult to use for planning or policy extraction. To address these limitations, we propose RoboTALES, a single-stage framework that learns task-aligned simulated futures and uses them to train robot policies. Our approach introduces two key innovations: (1) a hierarchical LLM-based planner that breaks complex tasks into a sequence of subgoals to guide the model's imagination; and (2) a VLM-based critic that evaluates these ``imagined'' futures and uses reward-based feedback to keep the model's internal representations focused on the goal. By anchoring the video generator in abstract reasoning, we produce temporally consistent rollouts and more coherent actions. We evaluate RoboTALES on diverse manipulation tasks from RoboCasa and LIBERO10, and show that our method consistently outperforms existing methods, especially in long-horizon tasks. Our code and models are publicly available at \textcolor{magenta}{\url{https://github.com/hananshafi/RoboTALES}}. 

\end{abstract}
\section{Introduction}
\label{sec:intro}
Humans rarely act purely reactively; instead, they decompose goals into ordered subtasks, mentally simulate possible futures, and commit to an action only after rejecting undesirable outcomes~\citep{miller1960plans, botvinick2008hierarchical}.
This capacity for hierarchical, goal-directed imagination remains an ongoing challenge for autonomous agents. Agents operating on high-dimensional visual observations must reason over extended horizons, anticipate the consequences of their actions, and select among many plausible futures. While model-free reinforcement learning can achieve strong task performance~\citep{mnih2015human, schulman2017proximal, haarnoja2018soft}, it offers no explicit reasoning substrate and remains sample-inefficient. Model-based approaches address this by learning a world model that enables an agent to ``imagine before acting''~\citep{sutton1991dyna, ha2018worldmodels, hafner2019planet, schrittwieser2020mastering, hafner2023dreamerv3, hansen2024tdmpc2}.

The idea of leveraging diffusion-based~\citep{ho2020ddpm,song2020denoising,blattmann2023stable}  video generation models~\citep{sora,chandra2025diwa} directly for robot control has recently gained considerable momentum. Works such as Video-Policy~\citep{liang2025videoGenerators}, Gen2Act~\citep{bharadhwaj2024gen2act}, and ViPRA~\citep{routray2025vipra} demonstrate that video generators can serve as powerful priors for policy learning, using generated rollouts to either supervise or directly condition action generation.  However, video generation alone does not yet constitute a world model for control. This is because the underlying video generators are still optimized primarily for visual realism or reconstruction objectives, hence their simulated futures can remain weakly grounded in task intent; moreover, making generated videos faithfully follow robot actions is itself nontrivial~\citep{li2025worldevalworldmodelrealworld}. 

In parallel, LLMs have demonstrated a remarkable ability to decompose long-horizon tasks into grounded subtasks and guide robot behavior through natural language~\citep{ahn2022saycan, huang2022innermonologue, brohan2023rt2, shah2023lmnav}. Yet, most prior works treat language and prediction as loosely coupled modules: language influences \emph{which} action to execute, but does not directly shape the predictive representations inside the world model. This decoupling prevents closed-loop alignment between what the agent imagines and what it ultimately does. Recent generative video world models further illustrate the promise of learned rollouts for control~\citep{bruce2024genie, hu2023gaia1}, but their integration with language-grounded reasoning and policy learning remains largely unexplored. Consequently, there is no closed-loop mechanism ensuring that imagined futures, high-level reasoning, and low-level actions remain semantically aligned.

These motivate a mechanism that can explicitly \emph{(i)} impose hierarchical structure on imagination and \emph{(ii)} enforce semantic faithfulness of simulated futures---both of which are crucial if imagined rollouts are to reliably benefit robot action generation. To address this gap, we introduce \textbf{RoboTALES}: Learning Reasoning-Guided {\textbf{Robo}t Policies via {\textbf{T}ask-{\textbf{AL}igned Simulat{\textbf{E}d Future\textbf{S}, a unified framework that formalizes the alignment between high-level task logic and low-level visual dynamics. As illustrated in Figure~\ref{fig:teaser}, RoboTALES operates through two tightly coupled components. An \textbf{LLM Planner } that decomposes the natural task instruction into an ordered sequence of temporal subtasks. This hierarchical task decomposition, as studied in both cognitive science~\citep{botvinick2008hierarchical, miller1960plans} and machine learning~\citep{pateria2021hierarchical, andreas2017modular}, provides an explicit temporal scaffold, analogous to human subgoal reasoning~\citep{botvinick2008hierarchical}. Conditioning the video diffusion model on these subtasks transforms rollout generation from an undifferentiated prediction problem into a structured, milestone-driven simulation process. 
However, generating structured rollouts is necessary but not sufficient. Even with planner conditioning, a generative model may produce futures that satisfy low-level visual statistics while violating high-level intent. Therefore, to enforce semantic faithfulness, we introduce a VLM-based \textbf{Critic} that closes this gap by evaluating the generated rollouts against the target task instruction, producing a language-conditioned alignment score. Crucially, this feedback is incorporated in the video generator's hidden states during training via a differentiable policy optimization~\citep{black2023ddpo}.
This representational steering refines the internal dynamics of the diffusion VideoUNet, encouraging it to encode task-relevant semantics rather than purely visual statistics. These semantically enriched features then serve as the foundation for generating 
purposeful, precise motor actions.

The joint integration of planner and critic transforms the diffusion VideoUNet into a goal-conditioned transition function where the planner constrains which futures are imagined, and the critic ensures those futures are worth acting upon. We show that our goal-aware features provide a much stronger foundation for robot control. By linking the robot's actions directly to these reasoned `imaginations', the policy produces smoother and more purposeful movements. This effectively eliminates the erratic behaviors common in traditional diffusion policies~\citep{chi2024diffusionpolicy}, ensuring the robot stays on track even during long, complex tasks. 
Empirically, RoboTALES improves long-horizon manipulation on RoboCasa and LIBERO. In particular, on challenging Pick-and-Place tasks in RoboCasa, RoboTALES reaches a mean success rate of 48\%, outperforming \cite{liang2025videoGenerators} and the strongest prior baseline \cite{han2024dualprocessvlaefficient}. On multi-step tasks involving turning and pressing, RoboTALES achieves mean success rates of 64\% and 96\%, respectively, surpassing competing methods.
Overall, these results support our core thesis: \emph{aligning imagination with hierarchical task logic yields more reliable rollouts and more coherent downstream actions}.

\noindent In summary, our contributions are:
\begin{itemize}
    \item \textit{Planner-Conditioned Video Generation.}
    We integrate hierarchical subtask plans from an LLM planner into the latent space of a diffusion-based video generator model. This induces structured, temporally organized rollouts that mirror abstract, goal-directed reasoning.

    \item \textit{Critic-Guided Representational Steering.}
    We leverage VLM-based reward signal within an RL-based differentiable optimization framework enabling language-aligned feedback to directly refine the video generator's latent dynamics. This ensures semantic faithfulness to the task instruction.

    \item \textit{Reasoning-Aligned Policy Learning.} We show that conditioning action generation on reasoning-aligned predictive features produces more precise and semantically consistent trajectories.

\end{itemize}

\section{Related Work}
\label{sec:related-work}

\textbf{Diffusion Policy Models.}
Due to over-conservatism, limited expressiveness, and compounding errors in offline policy learning and planning, a growing body of work adopts diffusion models as policy parameterizations. Diffuser~\cite{janner2022diffuser} first proposed trajectory-level diffusion to jointly predict all timesteps of a plan, improving temporal consistency and reducing compounding errors. Subsequent methods~\cite{wang2023diffusioninrl,chen2023offline,he2023diffcps,10423845,hansenestruch2023idql} represent policies as diffusion models to capture multimodal action distributions and increase expressiveness. Diffusion Policy~\cite{chi2024diffusionpolicy} brought this paradigm to robotics, with follow-ups extending to 3D settings~\cite{3d_diffuser_actor,ze20243d,yan2024dnact}, scaling~\cite{zhu2024scaling}, improving efficiency~\cite{pmlr-v270-jia25b,wang2024onedp}, and introducing architectural refinements~\cite{pmlr-v270-zhao25b,wang2024sparse,prasad2024consistency,reuss2024multimodal}. 
Despite strong empirical performance, these approaches are primarily reactive and do not explicitly endow the policy with a reasoning substrate: they learn to map observations to actions, but do not train an internal predictive model whose rollouts are structured by plans or validated for semantic faithfulness. In contrast, our RoboTALES learns robot policies from a reasoning-guided diffusion predictive model, where planning and critique actively shape the representations used for control.

\medskip
\noindent
\textbf{Video Generation for Robot Actions.}
A rapidly growing body of work leverages large-scale video generative models as world 
models for robot control~\cite{sora, unisim, nvidia2025cosmosworldfoundationmodel, genie2}, 
motivated by the observation that video prediction implicitly captures physical dynamics, 
object interactions, and causal structure that can be repurposed for policy 
learning~\cite{hu2023gaia1, bruce2024genie}. VideoPolicy~\cite{liang2025videoGenerators} builds on Stable 
Video Diffusion (SVD)~\cite{blattmann2023stable} and conditions a 1D action UNet on hidden embeddings extracted from the video UNet's decoder 
layers, demonstrating strong generalization to unseen objects, backgrounds, and tasks. 
Gen2Act~\cite{bharadhwaj2024gen2act} similarly synthesizes video demonstrations of unseen tasks to guide 
policy learning, while ViPRA~\cite{routray2025vipra} uses video predictions as affordance signals for 
action generation.
These works establish video generation as a powerful proxy for policy learning and form the 
direct foundation that our proposed RoboTALES builds upon.

\medskip
\noindent\textbf{Planning via Language.}
LLMs and VLMs have shown strong capacity for task decomposition and high-level reasoning, motivating their use as planning modules for embodied agents. \cite{ahn2022saycan} grounds LLM plans in robot affordances, \cite{huang2022innermonologue} incorporates environment feedback into an LLM reasoning loop for closed-loop replanning, and \cite{shah2023lmnav} decomposes navigation instructions into visual waypoints using LLMs and VLMs. More recently, \cite{brohan2023rt2} embeds reasoning directly into action prediction by fine-tuning VLMs end-to-end on robot demonstrations, while \cite{chen2025planningreasoningusingvision} performs high-level planning entirely in language space by predicting interleaved actions and state descriptions. While these approaches demonstrate that language is a powerful medium for abstract task reasoning, they share a common limitation: language reasoning remains decoupled from the agent's predictive representations. Plans influence which actions are selected but do not shape what the agent imagines about the future. In RoboTALES, the Planner conditions the generative process itself, embedding structured task reasoning into the latent visual dynamics rather than treating it as an external directive.

\section{RoboTALES}
\label{methodology}

\subsection{Problem Setup and Formulation}
\label{subsec:problem_formulation}
We consider an embodied agent tasked with completing a long-horizon goal described by a natural-language instruction $\tau$. At each time step $t$, the agent receives a visual observation $s_t \in \mathcal{S}$ and must produce a sequence of continuous actions $\mathbf{a}_{t:t+H} \in \mathcal{A}^{H}$ that maximizes task progress, where $H$ is the action prediction horizon. 

Our framework decomposes this problem into a hierarchical decision-making process. First, a high-level planner $\mathcal{F}_{P}$ maps the goal instruction $\tau$ to a structured plan consisting of semantically coherent sub-goals. Second, a generative world model $\mathcal{G}_{\theta}$ predicts a sequence of future latent states $\hat{s}_{t:t+\Delta}$ conditioned on the current state $s_t$ and the plan. A reward function $\mathcal{F}_{R}(\hat{s}, \tau)$ evaluates these imagined futures to provide a semantic alignment signal. Finally, an action policy $\pi_{\phi}$ maps the internal representations of the model to executable control signals $\mathbf{a}_{t:t+H}$. The objective is to jointly optimize the world model and the policy such that the generated actions align with the high-level reasoning provided by the planner.

A natural question is whether to train these components separately or jointly. Decoupled (two-stage) training---where the video generator is first trained in isolation and the action policy is subsequently trained on frozen features---avoids gradient interference between heterogeneous objectives but introduces a fundamental limitation: the video model has no awareness of which latent features are most useful for control, and the action policy must adapt to a fixed representation that was never optimized for action generation. In contrast, RoboTALES adopts a coupled, single-stage training strategy in which the video generator and action policy are optimized simultaneously. This end-to-end design allows action-level gradients to flow back into the video generator's decoder layers, encouraging it to produce latent representations that are not only visually faithful and semantically aligned but also directly informative for downstream control. The result is a tighter co-adaptation between imagination and action: the video model learns to ``imagine for acting,'' while the policy learns to ``act from imagination.''

\begin{figure}[t]
    \centering
    \includegraphics[width=1.0\textwidth]{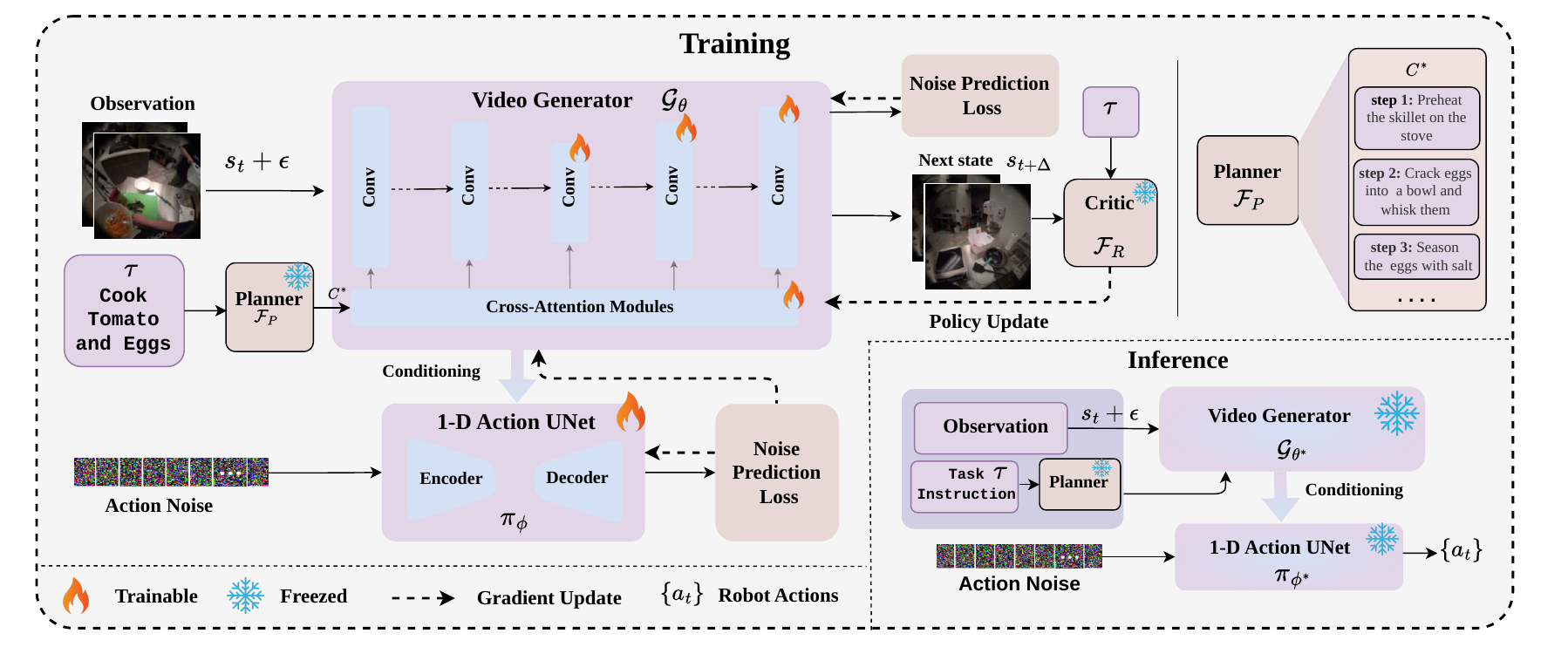}
    \caption{\small \textbf{RoboTALES Architecture and Training.} The Planner $\mathcal{F}_{P}$ decomposes the task 
instruction $\tau$ into ordered subtasks, forming the augmented 
plan $C^*$. The video generator $\mathcal{G}_\theta$ is jointly trained with 
the Action UNet $\pi_\phi$ in a single stage. $\mathcal{G}_\theta$ receives a 
video diffusion loss and reward feedback from a frozen VLM Critic 
$\mathcal{F}_{\mathcal{R}}$, while $\pi_\phi$ is trained with an action 
diffusion loss conditioned on the decoder features of $\mathcal{G}_\theta$. 
Crucially, action-level gradients propagate back into the selected 
decoder layers of $\mathcal{G}_\theta$, enabling the world model to co-adapt 
its representations for both visual prediction and downstream control. 
\textit{Inference (bottom-right):} Given a new observation and task 
instruction, the Planner produces $C^*$, the adapted world model 
$\mathcal{G}_{\theta^*}$ generates future states, and the trained policy 
$\pi_{\phi^*}$ outputs executable robot actions $\{a_t\}$ from the 
co-adapted latent representations.}
    \label{fig:arch}
    \vspace{-2em}
\end{figure}

\subsection{Architecture and Components}
\label{sec:arch}
RoboTALES consists of four components (Fig.~\ref{fig:arch}): a reasoning-based LLM Planner ($\mathcal{F}_P$), a video generator ($\mathcal{G}_\theta$), an action generation model ($\pi_\phi$), and an RL-based VLM critic ($\mathcal{F}_\mathcal{R}$). While these elements resemble hierarchical model-based RL architectures, our contribution lies in the interfaces between them, which enable task-aligned imagination. Specifically, the planner produces semantic subgoal tokens that condition the diffusion model, guiding the generation of future visual states. The imagined trajectories are evaluated by a vision-language critic that assesses their semantic alignment with the task description. Crucially, the action policy is conditioned on hierarchical representations extracted from the predicted trajectories, and its training signal flows back to shape these representations. This coupled design allows language reasoning, generative prediction, and control to co-evolve during training, enabling the agent to imagine, evaluate, and act upon task-consistent futures. We next describe each component.

\noindent
\textbf{Planner $\mathcal{F}_{P}$.} We instantiate $\mathcal{F}_P$ with a strong LLM \cite{comanici2025gemini} to serve as the high-level reasoning engine. The planner decomposes a long-horizon task instruction $\tau$ into a sequence of $K$ semantically coherent sub-tasks $C = \{c^{(1)}, \dots, c^{(K)}\}$ based on the current observation such that $c^{(1:K)} \sim \mathcal{F}_{P}(\cdot \mid s_t, \tau)$. This converts a compressed instruction into an explicit milestone sequence, providing dense semantic structure that improves long-horizon coherence compared to using $\tau$ alone~\cite{chen2025vlwm}.
In practice, we use $K\in[2,5]$ depending on task complexity.

\noindent
\textbf{Video Generator $\mathcal{G}_{\theta}$.} We adopt the pretrained Stable Video Diffusion (SVD) backbone \cite{blattmann2023stable} as the next state generator.
Conditioned on the visual state $s_t$ and the planner output, the model predicts a short-horizon future rollout of $\Delta$ frames represented as $
    \hat{s}_{t+1:t+\Delta}$.

\noindent
\textbf{VLM Critic $\mathcal{F}_{\mathcal{R}}$.} 
A frozen vision-language model, parameterized by $\psi$, evaluates the generated future rollouts $\hat{s}_{t+1:t+\Delta}$ against the goal instruction $\tau$. The critic operates in pixel space to produce a scalar reward signal $r$ that reflects semantic alignment, realism, and goal satisfaction:
This signal is used to optimize the targeted parameters of $\mathcal{G}_{\theta}$ via differentiable policy optimization \cite{black2023ddpo}, reinforcing denoising transitions that yield high-reward visual states.

\noindent
\textbf{Action Generator $\pi_{\phi}$.} 
The policy $\pi_{\phi}$ is implemented as a 1D diffusion UNet \cite{chi2024diffusionpolicy}. The action generator is conditioned on initial action noise $a_i$ (where $i$ denotes the denoising step) and the hidden state embeddings extracted from the video generator, and generates continuous executable action sequences $\mathbf{a}_{t:t+H}$.

\subsection{Planning with Hierarchical Reasoning}
A central insight of our framework is that raw task instructions are often too compressed and ambiguous to directly guide a generative model over long horizons. To bridge this semantic gap, we introduce an explicit planner that decomposes high-level goals into structured, actionable hierarchical sub-goals before any visual prediction or action generation takes place. In RoboTALES, however, the decomposition serves an additional role beyond simplifying control: the planner produces semantic subgoal tokens that condition the diffusion world model during generation. These tokens act as intermediate anchors that guide the imagined trajectory of future states, ensuring that predicted visual rollouts remain aligned with task intent. This integration enables task-conditioned visual imagination, where the video model predicts not only physically plausible futures but also futures that reflect the evolving semantic goals of the task.

Given an original task instruction $\tau$, the planner $\mathcal{F}_P$ generates a sequence of $K$ semantically coherent sub-goals: $c^{(1:K)} \sim \mathcal{F}_{P}(\cdot \mid s_t, \tau)$,
where each sub-goal $c^{(k)}$ describes a concrete, short-horizon objective grounded in the agent's current visual context.
For instance, a compressed instruction such as $\tau{=}$\,\texttt{"Prepare the ingredients for Zucchini Curry"} is decomposed into specific directives: $c^{(1)}{=}$\,\texttt{"Wash, peel, and chop the zucchini"}, $c^{(2)}{=}$\,\texttt{"Chop the onions and \linebreak tomatoes"}, and so forth. 
In practice, we extract $K \in [2, 5]$ sub-goals per instruction, scaling with the inherent complexity of the task. The sub-goals are then concatenated with the original instruction to form an augmented, reasoning-rich plan,
\begin{equation}
    C^* = \big[\,\tau\;;\; c^{(1)}\;;\; c^{(2)}\;;\; \dots \;;\; c^{(K)}\,\big].
    \label{eq:Cstar}
\end{equation}
Unlike the initial compressed instruction, $C^*$ provides a detailed roadmap that makes implicit task knowledge explicit. It provides a dense semantic signal to the video generator, enabling it to anticipate future states that are not merely visually plausible but also causally aligned with the task's logical progression.

\subsection{Task-Aligned Predictive Model Steering and Action Decoding}
\label{subsec:coupled-training}
 RoboTALES trains the video generator $\mathcal{G}_\theta$ and the action policy $\pi_\phi$ jointly in a single stage. This coupled optimization is motivated by a key observation: a predictive model becomes most useful for control when its internal representations are shaped not only by visual fidelity and semantic alignment, but also by the downstream action objective. In a decoupled regime, the video model's decoder features are fixed before the action policy ever sees them, meaning the model has no incentive to produce representations that are maximally informative for control. Joint training closes this loop by enabling the action-level gradients to flow back into the video generator's decoder layers, encouraging them to encode features that are simultaneously good for prediction and for acting.

During training, the planner $\mathcal{F}_P$ provides the augmented plan $C^*$, and we adapt the video diffusion UNet $\mathcal{G}_\theta$ while preserving its pretrained visual prior. Specifically, we freeze all parameters except \emph{(i)}~the cross-attention modules that ingest CLIP embeddings of $C^*$ and \emph{(ii)}~a targeted subset of decoder layers whose hidden states condition the action UNet. Let $\theta^* \subset \theta$ denote these trainable parameters. This targeted update constrains learning to the interfaces where task semantics enter the model (text cross-attention) and where task-relevant abstractions emerge (selected decoder blocks), enabling the video generator to become instruction-aware without sacrificing generative stability.

The coupled objective comprises three complementary losses: a video diffusion loss for visual fidelity, a policy optimization loss for semantic alignment, and an action diffusion loss for precise control.

\paragraph{Video Diffusion Loss.}
We optimize the selected parameters using the standard noise-prediction objective,
\begin{equation}
L_{\mathrm{video}}(\theta^*)
=\mathbb{E}_{x_0,C^*,t,\epsilon}\big\|\epsilon-\epsilon_{\mathcal{G}_{\theta^*}}(x_t,\sigma_t,C^*)\big\|^2,
\qquad
x_t=\alpha_t x_0+\sigma_t\epsilon,\;\epsilon\sim\mathcal{N}(0,\mathbf{I}).
\tag{2}
\end{equation}
\noindent
\textit{Task Alignment via VLM Critic.} While the diffusion loss encourages fidelity to the training distribution, it does not directly optimize for task-level semantic alignment.
To enforce this alignment, we incorporate a pretrained VLM-based critic $\mathcal{F}_\mathcal{R}$ that scores decoded future states against goal instruction $\tau$. We cast the iterative denoising process as a multi-step MDP and apply differentiable policy optimization~\cite{black2023ddpo}, where each denoising step constitutes an action taken by the video model. Let $x_t \in \mathcal{Z}$ be the latent state at noise level $\sigma_t$. The VideoUNet first predicts a clean latent estimate: $x_{0,\theta} = \mathcal{G}_{\theta^*}(x_t, \sigma_t, C^*)$.
Using Euler--Ancestral updates, the transition mean to the next less noisy state becomes,
\begin{equation}
\mu_\theta(x_t,\sigma_t,C^*)
=x_t+(\sigma_{t-1}-\sigma_t)\frac{x_t-x_{0,\theta}}{\sigma_t}.
\tag{3}
\end{equation}
We treat each sampler step as a Gaussian transition,
\begin{equation}
p_{\mathcal{G}_\theta^*}(x_{t-1}\!\mid x_t,C^*)
=\mathcal{N}\!\big(\mu_\theta(x_t,\sigma_t,C^*),\Sigma_t\big),
\quad
\Sigma_t=(\sigma_t^{\mathrm{ker}})^2\mathbf{I},
\tag{4}
\end{equation}
where $\sigma_t^{\mathrm{ker}} = \eta\sqrt{\sigma_t^2 - \sigma_{t-1}^2}$ controls the stochasticity of the sampler and $\eta$ is a tunable noise coefficient.

\noindent
\textit{Per-step Log-Likelihood.}
For a realized transition in a $d$-dimensional latent space, the log-likelihood of transitioning from $x_t$ to $x_{t-1}$ is,
\begin{equation}
\ell_\theta(x_{t-1}\!\mid x_t,C^*)
=
-\frac{\|x_{t-1}-\mu_\theta(x_t,\sigma_t,C^*)\|_2^2}{2(\sigma_t^{\mathrm{ker}})^2}
-\frac{d}{2}\log\!\big(2\pi(\sigma_t^{\mathrm{ker}})^2\big).
\tag{5}
\end{equation}

\noindent
\textit{Reward Computation.} Given a complete denoising rollout $(x_T, \ldots, x_0)$, we decode selected keyframes into pixel-space observations represented as: $s_{t:t+\Delta}$ and evaluate them with the frozen VLM critic which gives a scaler reward $r$,
\begin{equation}
r=F_{\mathcal{R}}(s_{t:t+\Delta},\tau)\in\mathbb{R}.
\tag{6}
\end{equation}
We compute a per-instruction baseline $b(\tau)$ as a running mean of rewards for instruction $\tau$, yielding the advantage $A = r - b(\tau)$.

\noindent
\textit{Policy Optimization Objective.}
Following~\cite{black2023ddpo}, we optimize a REINFORCE-style objective over the denoising trajectory. Let $\mathcal{T}_K$ denote a uniformly sampled subset of $K$ denoising steps,
\begin{equation}
J_{\mathrm{DDPO}}(\theta^*)
=\mathbb{E}_{(x_T\dots x_0)}\!\left[
A\cdot\frac{1}{K}\sum_{t\in\mathcal{T}_K}\overline{\ell_\theta(x_{t-1}\!\mid x_t,\tau)}
\right],~
L_{\mathrm{DDPO}}(\theta^*)=-J_{\mathrm{DDPO}}(\theta^*).
\tag{7}
\end{equation}

\noindent
\textit{Action Diffusion Loss.}
Simultaneously, the action UNet $\pi_\phi$ receives initial action noise $\mathbf{a}_i$ (where $i$ denotes the denoising step) and hidden state embeddings $f_{\mathcal{G}_{\theta^*}}^{\{l\}}$ from the selected decoder layers of the video generator, and produces executable actions represented as $a_t \sim \pi_{\phi}\,\big(.\,|\, \mathbf{a}_i, f_{\mathcal{G}_{\theta^*},C^*}^{\{l\}}\big)$.
The action UNet is trained using the standard noise prediction objective,
\begin{align}
L_{\mathrm{action}}(\phi, \theta^*)
&=
\mathbb{E}_{(a,\varepsilon,i)}
\Big[
  \big\lVert
    \varepsilon
    -
    \pi_{\phi}\!\left(
      \cdot \,\middle|\,
      \mathbf{a}_i,\,
      f_{\mathcal{G}_{\theta^*},C^*}^{\{l\}}
    \right)
  \big\rVert^{2}
\Big]. \tag{8}
\label{eq:action_loss}
\end{align}
Critically, unlike decoupled approaches, we do not place a stop-gradient between $\pi_\phi$ and $\mathcal{G}_{\theta^*}$. Gradients from $L_{\mathrm{action}}$ propagate back through the conditioning features $f_{\mathcal{G}_{\theta^*}}^{\{l\}}$ into the video generator's trainable decoder layers $\theta^*$. This means the video model receives a direct training signal from the action objective; its decoder features are shaped not only to produce visually faithful and semantically aligned futures, but also to be maximally informative for action prediction. The video generator thus learns to ``imagine for acting''---encoding in its latent space the spatial, temporal, and semantic structure that the action policy needs for precise control.

\noindent
\textit{Joint Objective.}
The full training loss combines all three objectives,
\begin{equation}
L_{\mathrm{total}}(\theta^*, \phi)
= L_{\mathrm{video}}(\theta^*) + \beta\, L_{\mathrm{DDPO}}(\theta^*) + \gamma\, L_{\mathrm{action}}(\phi, \theta^*),
\tag{9}
\end{equation}
where $\beta$ and $\gamma$ are scaler weights balancing the generative, semantic, and control objectives.

This coupled formulation offers three advantages over decoupled training: \emph{(i)}~the video generator's decoder features co-adapt to the action objective, producing representations that encode control-relevant information that a purely generative loss would not incentivize; \emph{(ii)}~the action policy benefits from features that are continuously refined by both semantic and visual losses, rather than being constrained to a frozen representation that was never optimized for control; and \emph{(iii)}~the shared gradient flow creates an implicit feedback loop where improvements in action prediction inform better imagination, and better imagination in turn yields more informative features for action prediction. We validate this design choice through an ablation in the Appendix.

\section{Experimentation}
\noindent
\textbf{Implementation Details.}
\label{subsec:implementation}
Our implementation is built upon the codebase and experimental settings of \cite{liang2025videoGenerators}. We utilize Gemini-2.5-Pro \cite{comanici2025gemini} as the high-level Planner $\mathcal{F}_P$. The framework was trained on the RoboCasa dataset \cite{nasiriany2024robocasalargescalesimulationeveryday} using two NVIDIA A100 80GB GPUs. Joint training required approximately 6--7 days to complete with a batch size of 1 per GPU. Hyperparameters, including the learning rate and optimizer settings, were kept consistent with those in \cite{liang2025videoGenerators}. 
The VideoUNet and the Action-UNet were initialized with the Stage~2 weights from the official repository of \cite{liang2025videoGenerators}. For the reward module, we employed the reward choices from \cite{black2023ddpo}, \cite{bahng2025cycle} and \cite{ma2023liv}, serving as the reward metric for DDPO. Rewards were computed on decoded keyframes every 4 steps and remained frozen throughout training.
Typically, the input state is a single frame with 3 views, and 8 future rollouts are obtained per view. 

\noindent
\textbf{Simulation Setup and Baselines.}
We evaluate our method on the RoboCasa \cite{nasiriany2024robocasalargescalesimulationeveryday} and LIBERO10 \cite{liu2023liberobenchmarkingknowledgetransfer} benchmarks, spanning a total of 34 manipulation tasks. RoboCasa provides a large-scale, realistic simulation environment for complex manipulation research. For each task, both benchmarks provide 50 human demonstrations. Demonstrations in RoboCasa were replayed at a resolution of $256 \times 256$ pixels, and LIBERO10 demonstrations were resized to match this resolution. 
The action space is defined as $\mathbf{a}_t \in \mathbb{R}^7$, comprising the 6-DoF gripper pose and a scalar for the gripper state (open/closed). For RoboCasa, we follow the evaluation protocol of \cite{liang2025videoGenerators}, executing 50 rollouts across five distinct scenes per task. We use the baselines reported in \cite{liang2025videoGenerators}---including the RoboCasa-trained versions of \cite{li2025unifiedvideoactionmodel} and ResNet- and CLIP-based variants of \cite{chi2024diffusionpolicy}. For LIBERO10, we adhere to the protocols defined in \cite{li2025unifiedvideoactionmodel, liang2025videoGenerators}. 

\begin{table*}[t]
\centering
\caption{Comparison to the state of the art on the validation set of the RoboCasa simulation benchmark. Success rates are computed over 50 rollouts per task. Our method uses 50 demonstrations.}
\label{tab:robocasa-results}
\resizebox{0.98\textwidth}{!}{%
\begin{tabular}{ll|ccccccccc|c}
\toprule
\textbf{Category} & \multicolumn{1}{c|}{\textbf{Task}} & \textbf{3DA}  & \textbf{DP3} & \shortstack{\textbf{DP-}\\\textbf{ResNet}} & \shortstack{\textbf{DP-}\\\textbf{CLIP}} & \textbf{GR00T} & \textbf{FPV}  & \shortstack{\textbf{DP-}\\\textbf{VLA}}  & \textbf{UVA} & \shortstack{\textbf{Video-}\\\textbf{Policy}} & \textbf{Ours} \\
\midrule
\multirow{8}{*}{\shortstack[l]{\textbf{Pick and}\\\textbf{Place}}}
 & PnPCabToCounter        & 0.00 & 0.04 & 0.06 & 0.00 & 0.20 & 0.10 & 0.10 & 0.26 &0.16 & \textbf{0.44} \\
 & PnPCounterToCab        & 0.00 & 0.02 & 0.06 & 0.02 & 0.36 & 0.14 & 0.32 & 0.18 &0.38  & \textbf{0.44} \\
 & PnPCounterToMicrowave  & 0.00 & 0.06 & 0.06 & 0.02 & 0.13 & 0.10 & \textbf{0.56} & 0.10 &0.40  & 0.44 \\
 & PnPCounterToSink       & 0.00 & 0.00 & 0.14 & 0.08 & 0.10 & 0.08 & 0.30 & 0.16 &0.32  & \textbf{0.50}\\
 & PnPCounterToStove      & 0.00 & 0.00 & 0.00 & 0.02 & 0.24 & 0.04 & 0.22 & 0.16 & 0.52 & \textbf{0.54}\\
 & PnPMicrowaveToCounter  & 0.00 & 0.06 & 0.06 & 0.06 & 0.16 & 0.12 & 0.18 & 0.18 &0.16  & \textbf{0.26} \\
 & PnPSinkToCounter       & 0.00 & 0.00 & 0.10 & 0.22 & 0.33 & 0.30 & 0.56 & 0.38 &0.52  & \textbf{0.56} \\
 & PnPStoveToCounter      & 0.00 & 0.00 & 0.02 & 0.06 & 0.29 & 0.26 & 0.62 & 0.24 & 0.64 & \textbf{0.66}\\
\midrule
\multirow{4}{*}{\textbf{Doors}}
 & OpenSingleDoor  & 0.00 & 0.24 & 0.42 & 0.32 & 0.59 & 0.74 & 0.42 & 0.54 &0.78  & \textbf{0.80}\\
 & OpenDoubleDoor  & 0.00 & 0.20 & 0.70 & 0.82 & 0.15 & 0.92 & 0.80 & 0.90 &0.92  & \textbf{0.94} \\
 & CloseDoubleDoor & 0.00 & 0.56 & 0.78 & 0.84 & 0.75 & 0.78 & 0.84 & 0.76 & 0.90 & \textbf{0.94} \\
 & CloseSingleDoor & 0.14 & 0.62 & 0.78 & 0.48 & 0.83 & 0.84 & \textbf{1.00} & 0.88 & 0.98 & 0.98 \\
\midrule
\multirow{2}{*}{\textbf{Drawers}}
 & OpenDrawer  & 0.00 & 0.36 & 0.64 & 0.60 & 0.79 & 0.72 & 0.66 & 0.28 &0.40  & \textbf{0.8} \\
 & CloseDrawer & 0.00 & 0.48 & 0.82 & 0.96 & 0.99 & 0.94 & 1.00 & 0.72 &0.88 & \textbf{1.0} \\
\midrule
\multirow{2}{*}{\shortstack[l]{\textbf{Twisting}\\\textbf{Knobs}}}
 & TurnOnStove  & 0.10 & 0.24 & 0.38 & 0.28 & 0.56 & \textbf{0.66} & 0.64 & 0.50 &0.34 & 0.40\\
 & TurnOffStove & 0.02 & 0.06 & 0.16 & 0.08 & \textbf{0.27} & 0.20 & 0.16 & 0.14 & 0.06 & 0.10\\
\midrule
\multirow{3}{*}{\shortstack[l]{\textbf{Turning}\\\textbf{Levers}}}
 & TurnOnSinkFaucet  & 0.06 & 0.32 & 0.66 & 0.66 & 0.63 & 0.70 & 0.56 & 0.62 & 0.80 & \textbf{0.84} \\
 & TurnOffSinkFaucet & 0.28 & 0.42 & 0.68 & 0.70 & 0.73 & 0.78 & 0.72 & 0.64 & 0.68  & \textbf{0.78}\\
 & TurnSinkSpout     & 0.26 & 0.54 & 0.62 & 0.26 & 0.53 & 0.52 & \textbf{0.90} & 0.64 &0.26  & 0.32\\
\midrule
\multirow{3}{*}{\shortstack[l]{\textbf{Pressing}\\\textbf{Buttons}}}
 & CoffeePressButton & 0.08 & 0.16 & 0.76 & 0.68 & 0.85 & 0.90 & 0.86 & 0.84 & 0.92 & \textbf{0.96} \\
 & TurnOnMicrowave   & 0.06 & 0.38 & 0.68 & 0.88 & 0.78 & 0.68 & 0.84 & 0.94 & 0.86 & \textbf{0.96} \\
 & TurnOffMicrowave  & 0.32 & 0.54 & 0.62 & \textbf{1.00} & 0.71 & 0.96 & 0.86 & 0.96 & 0.94  & 0.96 \\
\midrule
\multirow{2}{*}{\textbf{Insertion}}
 & CoffeeServeMug & 0.00 & 0.18 & 0.44 & 0.60 & 0.73 & 0.48 & 0.64 & \textbf{0.78} & 0.74  & 0.60 \\
 & CoffeeSetupMug & 0.00 & 0.04 & 0.10 & 0.12 & 0.23 & 0.16 & \textbf{0.30} & 0.20 & 0.18 & 0.20\\
\midrule
\multicolumn{2}{l|}{\textbf{Avg. Success}} & 0.06 & 0.23 & 0.41 & 0.43 & 0.50 & 0.51 & 0.57 & 0.50 & 0.575  & \textbf{0.64} \\
\bottomrule
\end{tabular}%
}
\end{table*}

\begin{table}[!t]
\centering
\caption{Average success rates for tasks in the LIBERO10 benchmark.}
\label{tab:LIBERO10}
\begin{tabular}{l|cccccccc}
\toprule
Model & DP-C & DP-T & OpenVLA & $\pi_0$ & $\pi_0$-FAST & UVA & VideoPolicy & Ours \\
\midrule
Avg.\ & 0.53 & 0.58 & 0.54 & 0.85 & 0.60 & 0.90 & 0.94 & \textbf{0.97} \textbf{} \\
\bottomrule
\end{tabular}
\vspace{-1.5em}
\end{table}

\subsection{Quantitative Results}
Tables~\ref{tab:robocasa-results} and ~\ref{tab:LIBERO10} summarize success rates across a diverse set of long-horizon manipulation tasks. Overall, on RoboCasa simulation dataset (Table ~\ref{tab:robocasa-results}), \textbf{RoboTALES} attains the highest average success and improves consistently across categories, indicating that its gains are not confined to a narrow subset of tasks. Notably, the largest margins appear on \emph{structure-sensitive} tasks (\emph{e.g.}, doors and drawers) where the agent must execute temporally ordered subgoals (approach $\rightarrow$ grasp $\rightarrow$ actuate $\rightarrow$ release) and small deviations compound into failure. We also observe decent improvements on \emph{long horizon} tasks  (\emph{e.g.}, Pick and Place) and \emph{contact-rich} behaviors (\emph{e.g.}, turning knobs/levers and insertion), suggesting that aligning simulated futures to the instruction yields predictive features that better disambiguate fine-grained action choices. Notably, \cite{han2024dualprocessvlaefficient} and \cite{gemini2025robotics} employ substantially more demonstrations for behavior cloning (300 demos), yet our method surpasses both using only 50 demonstrations, underscoring the sample efficiency. In contrast to prior video-based action extraction pipelines~\citep{liang2025videoGenerators,bharadhwaj2024gen2act,routray2025vipra}\footnote{Reproduced numbers have been verified via correspondence with the
VideoPolicy \cite{liang2025videoGenerators} authors; they acknowledge a discrepancy in performance
between their paper and released codebase. To ensure a fair comparison, we
use the exact same experimental setup for all experiments in this paper.}, 
RoboTALES explicitly enforces \emph{task-aligned imagination} providing reasoning-rich features for a more reliable conditioning signal for action diffusion, translating into higher task completion rates across the benchmark.
This consistent improvement across diverse tasks in the RoboCasa distinct benchmarks spanning 24 manipulation tasks confirms the generalizability of our approach. Similarly, in Table \ref{tab:LIBERO10}, our method achieves a mean success rate of 97\% across all 10 tasks in the LIBERO10 benchmark \cite{liu2023liberobenchmarkingknowledgetransfer}, outperforming all baselines by a substantial margin as shown in Table \ref{tab:LIBERO10}. Among prior approaches, VideoPolicy represents the strongest competitor with a success rate of 94\%, followed by UVA at 90\% and $\pi_0$ at 85\%, while standard diffusion policy variants (DP-C: 53\%, DP-T: 58\%) and OpenVLA (54\%) lag considerably further behind.  
\begin{figure}[t]
  \centering
    \includegraphics[width=1.0\textwidth]{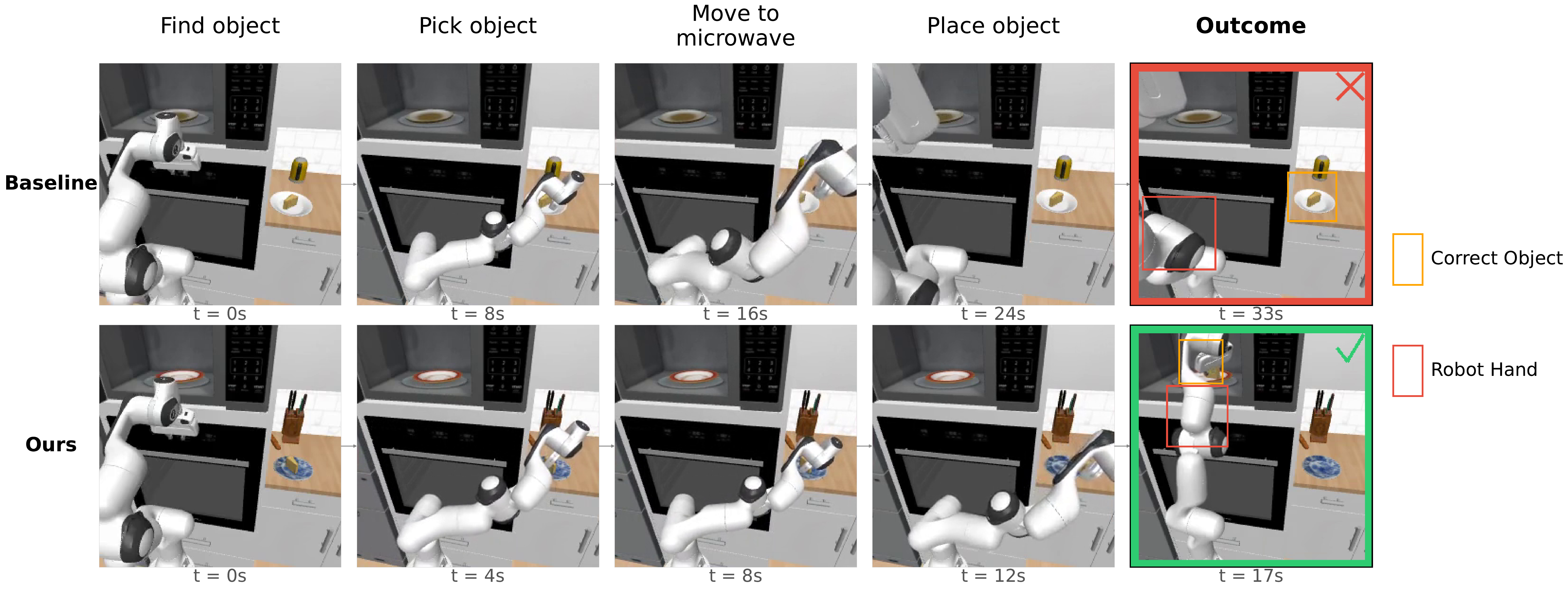}
    \caption{\textbf{Qualitative Comparison of Task Execution}. While the baseline model (top) fails to successfully pick up the target object and subsequently loses environmental consistency due to semantic drift, our framework (bottom) maintains structural integrity throughout the sequence and accurately executes the transition from picking the object to successfully placing it inside the microwave. \textit{(Best viewed zoomed in)}}
    \label{fig:qualitative-example}
    \vspace{-2em}
\end{figure}
\subsection{Qualitative Analysis}
\label{subsec:qualitative-analysis}
We conduct a qualitative comparison of our framework against the current state-of-the-art baseline, VideoPolicy \cite{liang2025videoGenerators}, focusing on task execution stability and semantic grounding. As illustrated in Fig. \ref{fig:qualitative-example}, we evaluate both models on manipulation tasks in the RoboCasa environment.
A primary failure in the baseline is semantic drift during long-horizon execution. While the baseline can initiate a grasp, it fails to maintain environmental integrity during complex transitions, such as moving an object toward a microwave. As seen in Fig. \ref{fig:qualitative-example} (top), the baseline exhibits physical distortions, leading to task failure.
In contrast, our framework maintains high fidelity. By conditioning the video model on reasoning-rich sub-goals the agent retains a strong semantic anchor. This prevents execution drift, ensuring trajectories remain physically plausible and goal-oriented.

\subsection{Ablations and Discussion}
In this section, we evaluate the individual contributions of our core components and discuss key properties.

\noindent
\textbf{Effect of the Planner. }
We compare three variants: \emph{(i)}~\cite{liang2025videoGenerators} augmented with our Planner at inference, \emph{(ii)}~our model trained with Planner conditioning but without the Critic, and \emph{(iii)}~our full model with both. As shown in Fig.~\ref{fig:planner_ablation} (averaged across 3 RoboCasa tasks), simply injecting structured sub-goals into an unadapted video model yields the lowest success rate, confirming that a pretrained model cannot exploit hierarchical plans without being explicitly trained to reason over them. Training video model with planner
 conditioning alone improves performance, confirming that the decomposed plan provides a stronger semantic signal than the raw instruction. Finally, our full method, which additionally incorporates VLM Critic feedback to align the video model's imagined futures with the task goal, achieves the highest success rate.
This demonstrates that the Planner is necessary but not sufficient on its own; its full benefit is realized when the video generator is trained to ground structured sub-goals into semantically coherent future predictions.
\begin{figure}[t]
  \centering
  \begin{minipage}{0.5\textwidth}
    \centering
    \includegraphics[width=\linewidth]{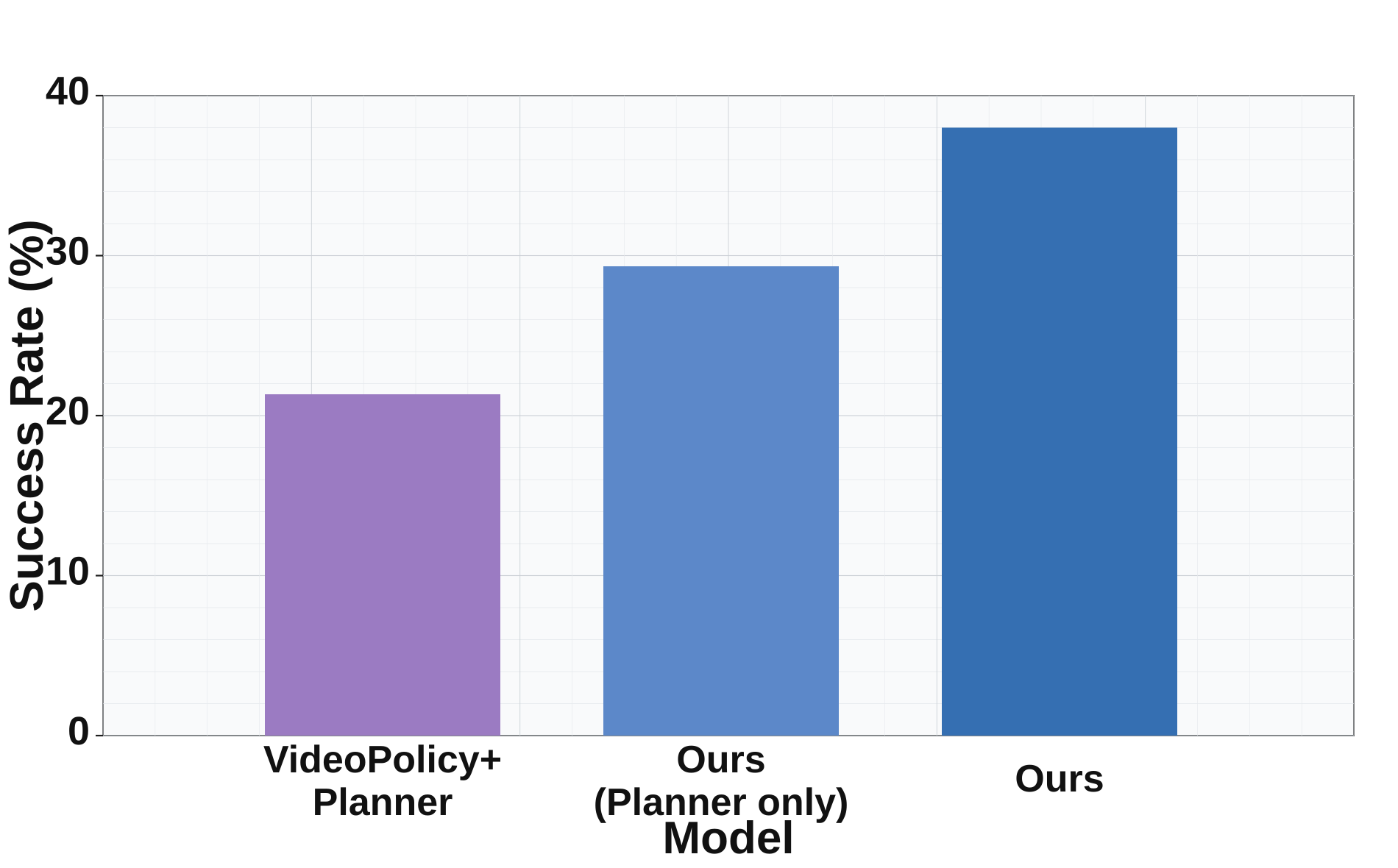}
  \end{minipage}%
  \begin{minipage}{0.5\textwidth}
    \raggedleft
\caption{\small \textbf{Impact of the Planner.} Planner provides structured subgoals, improving performance when trained with the video generator. Naively augmenting planner with baseline (purple graph) without adaptation results in performance degradation.}
\label{fig:planner_ablation}
  \end{minipage}
  \vspace{-2em}
\end{figure}

\begin{figure}[t]
  \centering
  \begin{minipage}{0.32\textwidth}
    \centering
    \includegraphics[width=\linewidth]{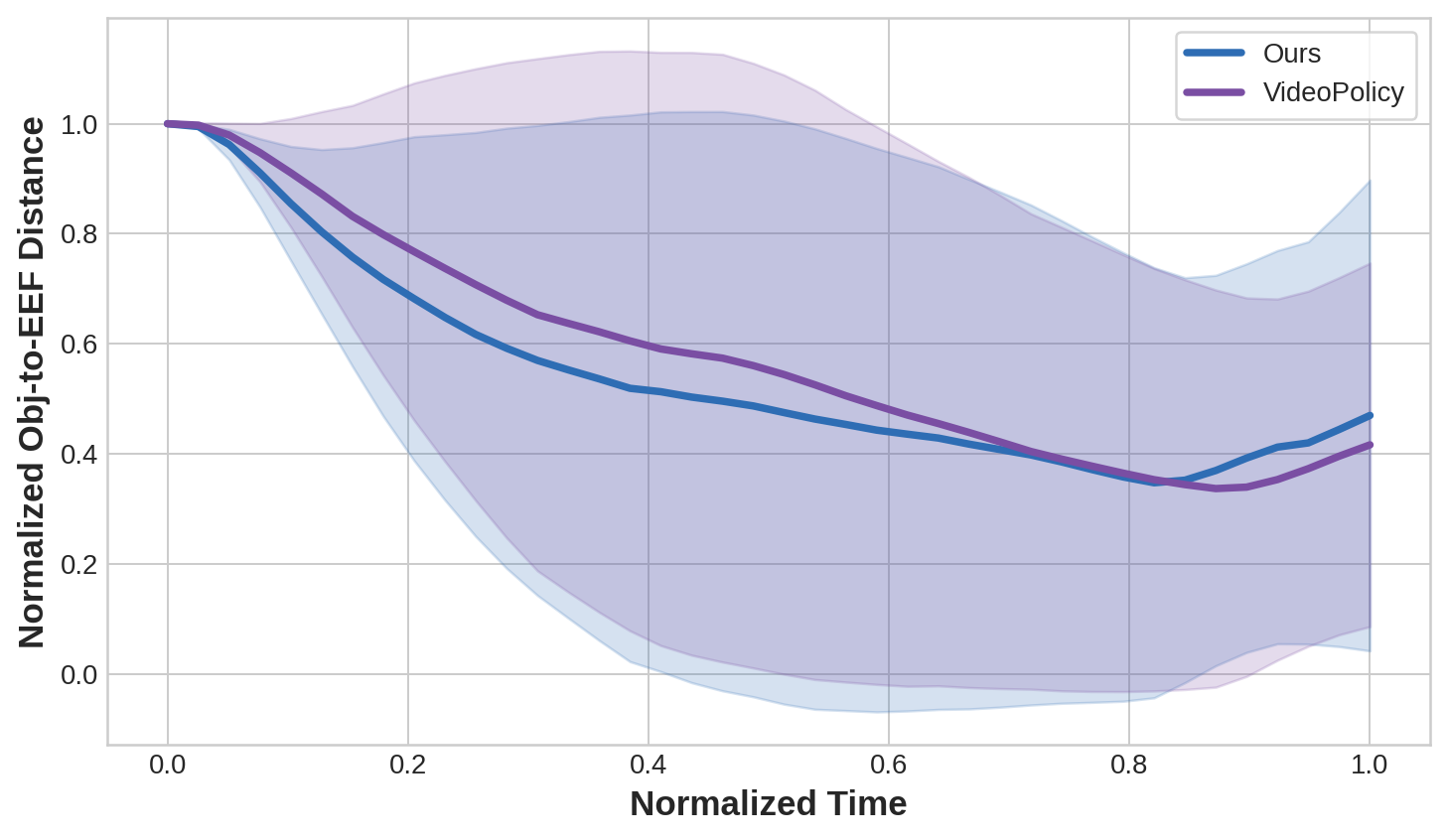}
    \subcaption{Trajectory Quality}
    \label{fig:robustness}
  \end{minipage}%
  \hfill
  \begin{minipage}{0.32\textwidth}
    \centering
    \includegraphics[width=\linewidth]{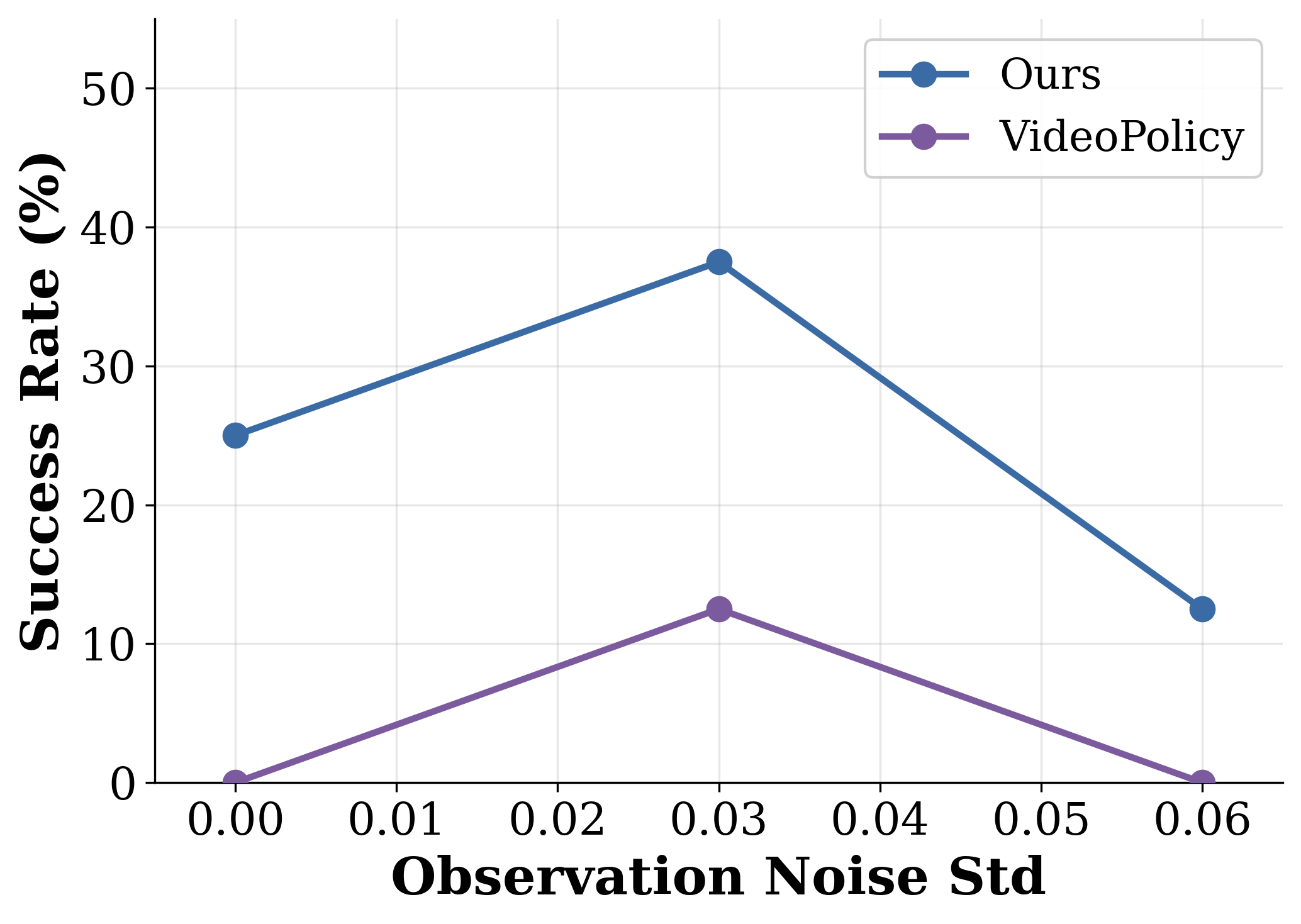}
    \subcaption{Robustness}
    \label{fig:trajectory}
  \end{minipage}%
  \hfill
  \begin{minipage}{0.3\textwidth}
    \centering
    \includegraphics[width=\linewidth]{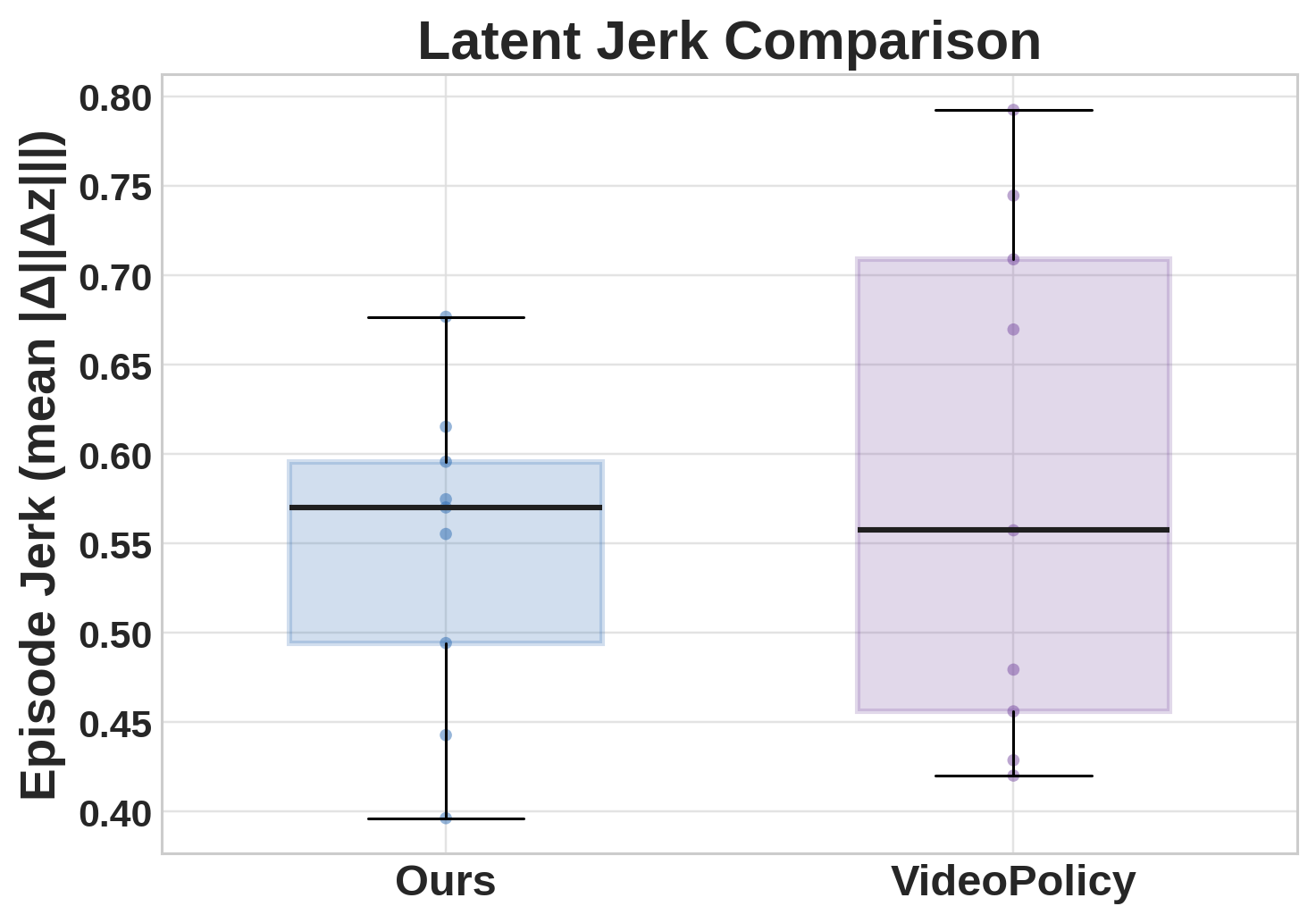}
    \subcaption{Latent jerk}
    \label{fig:placeholder}
  \end{minipage}
  \caption{\small \textbf{Analysis.}\textbf{(a)}~Normalized object-to-end-effector distance over time. Our method shows consistent purposeful behavior and lower variance,  \textbf{(b)}~Success rate (\%) under increasing 
  observation noise. Our method maintains significantly higher performance 
  across all noise levels, demonstrating robust semantic representations. 
  \textbf{(c)} RoboTALES produces smoother latent dynamics with lower jerk, indicating more temporally coherent representations that translate to smoother robot actions.}
  \label{fig:critic-analysis}
\end{figure}
\begin{figure}[!t]
  \centering
  \begin{minipage}{0.32\textwidth}
    \centering
    \includegraphics[width=\linewidth]{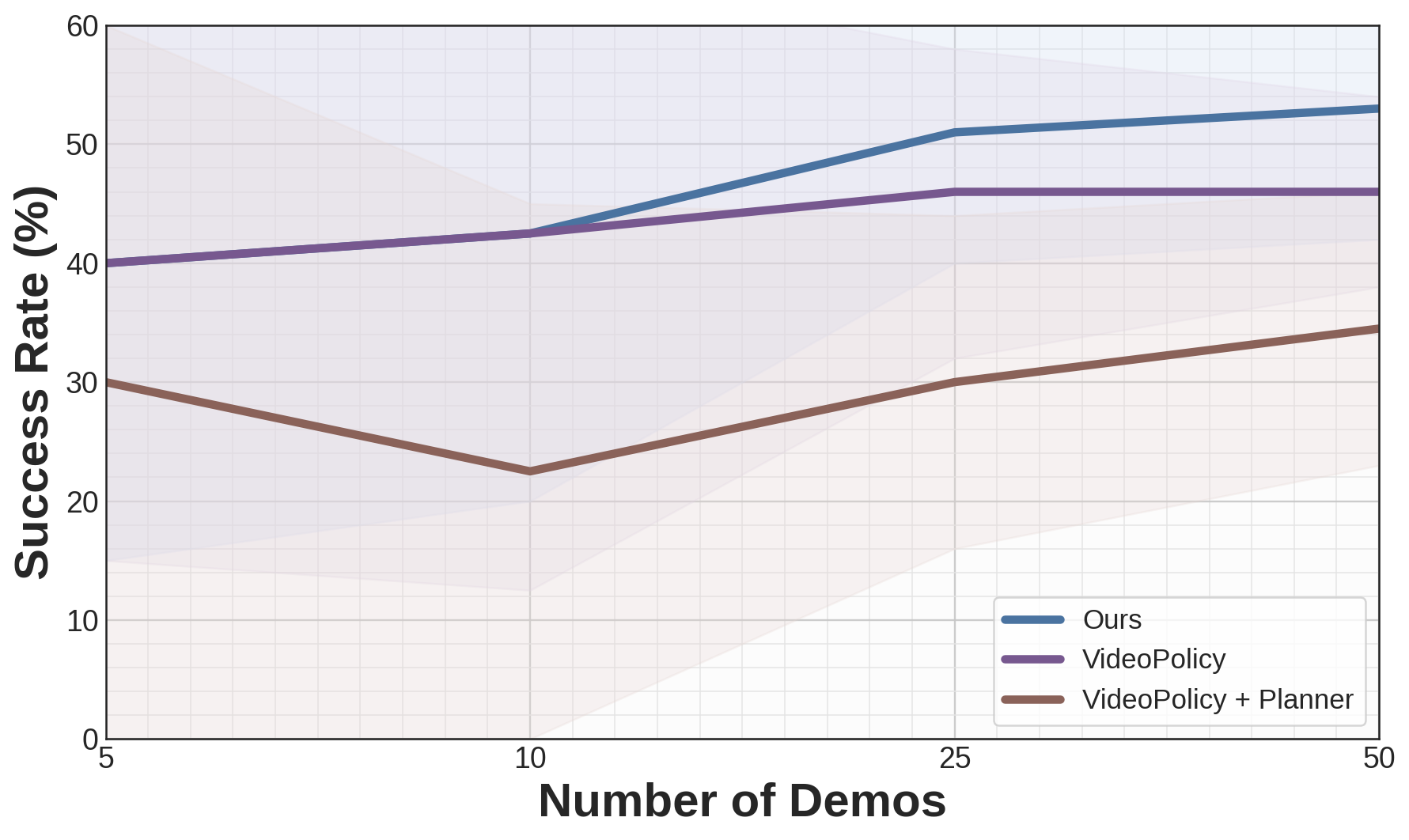}
    \subcaption{Demo Efficiency}
    \label{fig:demo-efficiency}
  \end{minipage}%
  \hfill
  \begin{minipage}{0.32\textwidth}
    \centering
    \includegraphics[width=\linewidth]{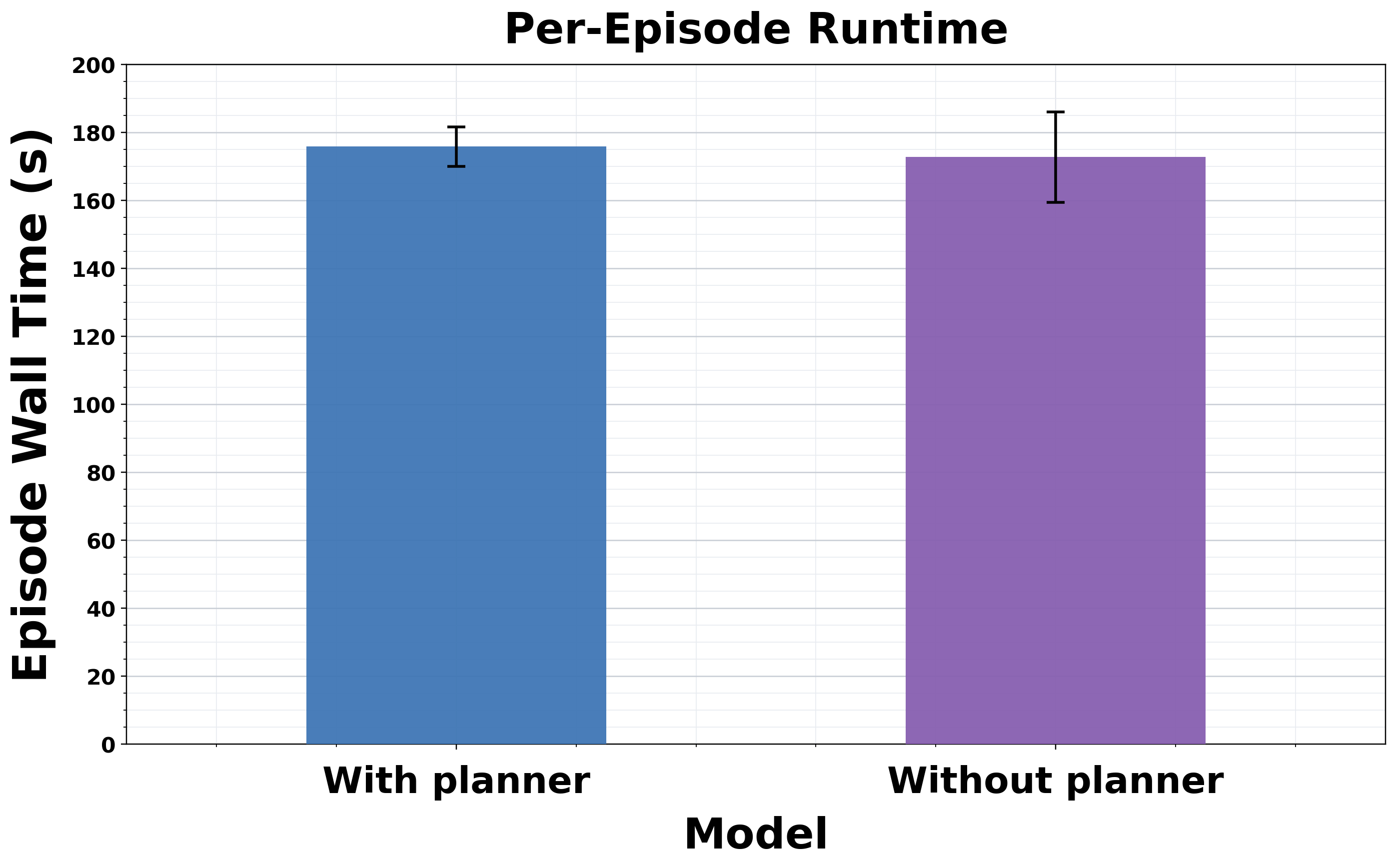}
    \subcaption{Inference Time}
    \label{fig:planner-time}
  \end{minipage}%
  \hfill
  \begin{minipage}{0.33\textwidth}
    \raggedright
    \small
    \textbf{Figure~\ref{fig:efficiency-analysis}:}
    \textbf{(a)}~Our method scales consistently with fewer demos, outperforming both baselines.
    \textbf{(b)}~The added overhead due to Planner is negligible ($\sim$1s), confirming execution efficiency.
  \end{minipage}
  \captionsetup{labelformat=empty}
  \caption{} 
  \label{fig:efficiency-analysis}
  \vspace{-4em}
\end{figure}

\noindent
\textbf{Effect of Policy Optimization.}
We examine whether RL-based refinement via the VLM Critic yields representations that improve downstream control beyond pixel-level prediction. We compare against \cite{liang2025videoGenerators}, which uses only a video diffusion loss with no semantic reward, and evaluate along three axes (Fig.~\ref{fig:critic-analysis}). In terms of \textit{trajectory quality} (Fig.~\ref{fig:critic-analysis}a), both methods progressively reduce the object-to-end-effector distance, but ours exhibits significantly lower variance, indicating more consistent reaching behavior from critic-aligned features. For \textit{robustness} (Fig.~\ref{fig:critic-analysis}b), we inject Gaussian noise $\sigma \in \{0.00, 0.03, 0.06\}$ into observations at inference; our method maintains substantially higher success rates across all noise levels while \cite{liang2025videoGenerators} collapses to near zero at $\sigma=0.06$. Finally, for \textit{action smoothness} (Fig.~\ref{fig:critic-analysis}c), we measure end-effector jerk as a proxy for motion quality, where our method produces lower jerk values reflecting smoother, more physically plausible actions. These results confirm that the Critic actively shapes the video model's internal representations to be more stable and robust for action generation, not merely improving visual fidelity.

\noindent
\textbf{Demonstration Efficiency. }
Fig.~\ref{fig:demo-efficiency} reports the demonstration efficiency of our method against two baselines, averaged over four RoboCasa tasks (PnPSinkToCounter, PnPStoveToCounter, PnPCounterToStove, PnPCounterToCab), across varying numbers of evaluation demonstrations. Even with as few as 10 demonstrations, our method achieves a competitive success rate of approximately 43\%, comparable to VideoPolicy \cite{liang2025videoGenerators} while significantly outperforming its planner-augmented variant. As the number of demonstrations increases, our method scales consistently, reaching approximately 51\% at 25 demonstrations and 53\% at 50, whereas VideoPolicy saturates around 46\% and its planner-augmented variant peaks at only 34\%. The shaded confidence intervals further reveal that our method exhibits reduced variance at higher demonstration counts, indicating more stable and reliable learning. 

\noindent
\textbf{Inference Time Comparison.}
A major concern with hierarchical models is the potential for slower performance due to multiple processing steps. Since our model only utilizes the Planner at the inference, therefore, the added cost is minimal. We compute the mean time taken by the model with and without Planner across 3 RoboCasa tasks spanning across 50 demos and 30 episodes. Fig. \ref{fig:planner-time} shows minimal increment in per-episode time at inference, which directly verifies the execution efficiency of our approach.

We refer reader to the Appendix for further results and discussions.

\section{Conclusion and Future Work}
\label{sec:conclusion}

We present RoboTALES, a framework that tightly couples hierarchical reasoning, video generation, and action generation for long-horizon robotic manipulation. By decomposing compressed task instructions into structured sub-goals via an LLM Planner and using them to condition a video model, our approach enables task-aligned visual imagination that goes beyond mere pixel-level fidelity. A frozen VLM Critic steers the video generator toward semantically faithful futures, producing latent representations that are simultaneously optimized for motor control. Extensive experiments across challenging benchmarks demonstrate that RoboTALES achieves impressive performance, outperforming strong baselines in task success rate, demonstration efficiency, long-horizon ability, and robustness to visual perturbations.

\noindent
\textit{Future Work.}
Our framework currently relies on a frozen VLM critic that provides a coarse semantic reward. A natural extension is to design task-aware reward functions that capture finer-grained progress signals such as sub-goal completion and physical plausibility. More ambitiously, we aim to explore learnable reward models that are jointly optimized with the video generator and the action policy, enabling the critic to co-adapt as the agent's capabilities improve.
%
%
\bibliographystyle{splncs04}
\bibliography{main}
\newpage
\appendix
\begin{center}
    {\textbf{RoboTALES: Learning Reasoning-Guided Robot Policies via Task-Aligned Simulated Futures}}
\end{center}
\section{\large Appendix}
The appendix is organized as below:
\begin{enumerate}[label=\textbf{A.\arabic*}, leftmargin=2em]
    \item Quantitative Results on LIBERO10 Benchmark (Sec.~\ref{appendix:additional-libero})
    \item Additional Ablations and Analysis (Sec.~\ref{appendix:additional-ablations})
    \item Additional Qualitative Results (Sec.~\ref{appendix:additional-qualitative})
    \item Planner Details (Sec.~\ref{appendix:planner_details})
    \item Limitations (Sec.~\ref{appendix:limitations})
\end{enumerate}

\subsection{Quantitative Results on Libero10 Benchmark}
\label{appendix:additional-libero}
Table~\ref{tab:LIBERO10} of the main paper demonstrated the mean success rate comparison across all 10 tasks in the LIBERO10 benchmark \cite{liu2023liberobenchmarkingknowledgetransfer}, where RoboTALES outperforms all baselines by a substantial margin 
The per-task breakdown in Table \ref{tab:libero10-pertask} reveals that our method achieves perfect or near-perfect success on the majority of tasks, including a perfect 1.0 on both "KITCHEN SCENE3 turn on the stove and put the moka pot on it" and "LIVING ROOM SCENE1 put both the alphabet soup and the cream cheese box in the basket," with only "KITCHEN SCENE8 put both moka pots on the stove" showing a relatively lower score of 0.88. Notably, the tasks where our model performs strongest tend to involve sequential multi-step manipulation requiring coherent long-horizon reasoning — precisely the setting where critic-guided semantic grounding and hierarchical planning provide the greatest benefit. The consistent high performance across diverse scene configurations (kitchen, living room, and study environments) further suggests that our reasoning-guided world model generalizes robustly rather than overfitting to a narrow set of visual contexts.


\begin{table}[h]
\centering
\caption{Per-task success rates out of 50 trials for the 10 tasks in the Libero10 benchmark, following the evaluation protocol in \cite{liang2025videoGenerators}.}
\resizebox{\linewidth}{!}{%
\begin{tabular}{l|c}
\toprule
\textbf{Libero10 Tasks} & \textbf{Success} \\
\midrule
LIVING ROOM SCENE2 put both the alphabet soup and the tomato sauce in the basket & 0.98  \\
LIVING ROOM SCENE2 put both the cream cheese box and the butter in the basket & 0.98 \\
KITCHEN SCENE3 turn on the stove and put the moka pot on it &  1.0\\
KITCHEN SCENE4 put the black bowl in the bottom drawer of the cabinet and close it &  0.96\\
LIVING ROOM SCENE5 put the white mug on the left plate and put the yellow and white mug on the right plate & 0.98 \\
STUDY SCENE1 pick up the book and place it in the back compartment of the caddy &  0.98\\
LIVING ROOM SCENE6 put the white mug on the plate and put the chocolate pudding to the right of the plate & 0.94 \\
LIVING ROOM SCENE1 put both the alphabet soup and the cream cheese box in the basket &  1.0\\
KITCHEN SCENE8 put both moka pots on the stove &  0.88\\
KITCHEN SCENE6 put the yellow and white mug in the microwave and close it & 0.98 \\
\midrule
\textbf{Average} & \textbf{0.97} \\
\bottomrule
\end{tabular}%
}
\label{tab:libero10-pertask}
\end{table}

\subsection{Additional Ablations and Analysis}
\label{appendix:additional-ablations}

\noindent
\textbf{Effect of Critic}
The critic assigns each rollout a scalar reward intended to predict task
success. To validate this signal, we evaluate it against ground-truth outcomes
over 360 rollouts (overall success rate $0.63$). Binning rollouts into reward
quartiles, success rate rises monotonically from $0.40$ in the lowest quartile
to $0.88$ in the highest (Fig.~\ref{fig:calibration}), confirming that the
critic is well calibrated: higher predicted reward reliably corresponds to a
higher chance of success. Treating the reward as a binary success classifier
yields an AUROC of $0.72$, well above the chance level of $0.5$.

How the per-step rewards are aggregated into a trajectory score matters. We
compare the \emph{mean} reward over the trajectory against the single-frame
\emph{max} (Fig.~\ref{fig:vlm-critic-auroc}). The mean aggregation achieves an
AUROC of $0.72$, substantially outperforming the max ($0.59$, near chance).
This indicates the critic's signal reflects sustained agreement across the
trajectory rather than a spurious peak at an individual frame, motivating our
use of the mean-aggregated reward.

\begin{figure}[t]
\centering

\begin{subfigure}{0.45\linewidth}
    \centering
    \includegraphics[width=\linewidth]{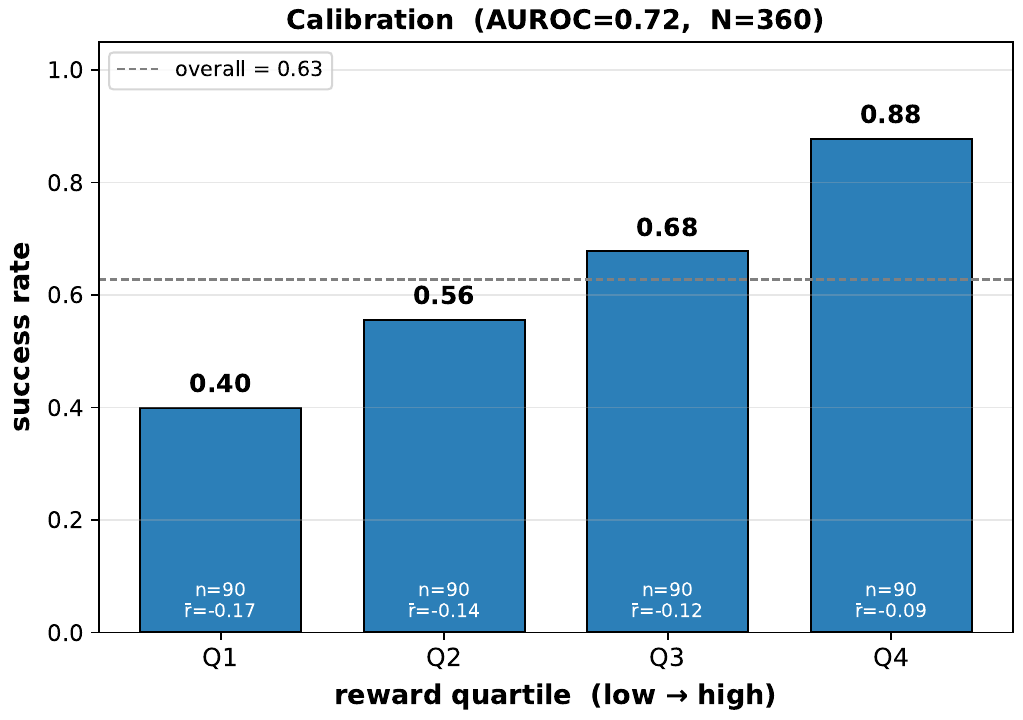}
    \caption{Success rate increases monotonically with the critic's predicted
    reward quartile (AUROC $=0.72$), showing the reward is well calibrated to
    task success.}
    \label{fig:calibration}
\end{subfigure}
\hfill
\begin{subfigure}{0.45\linewidth}
    \centering
    \includegraphics[width=\linewidth]{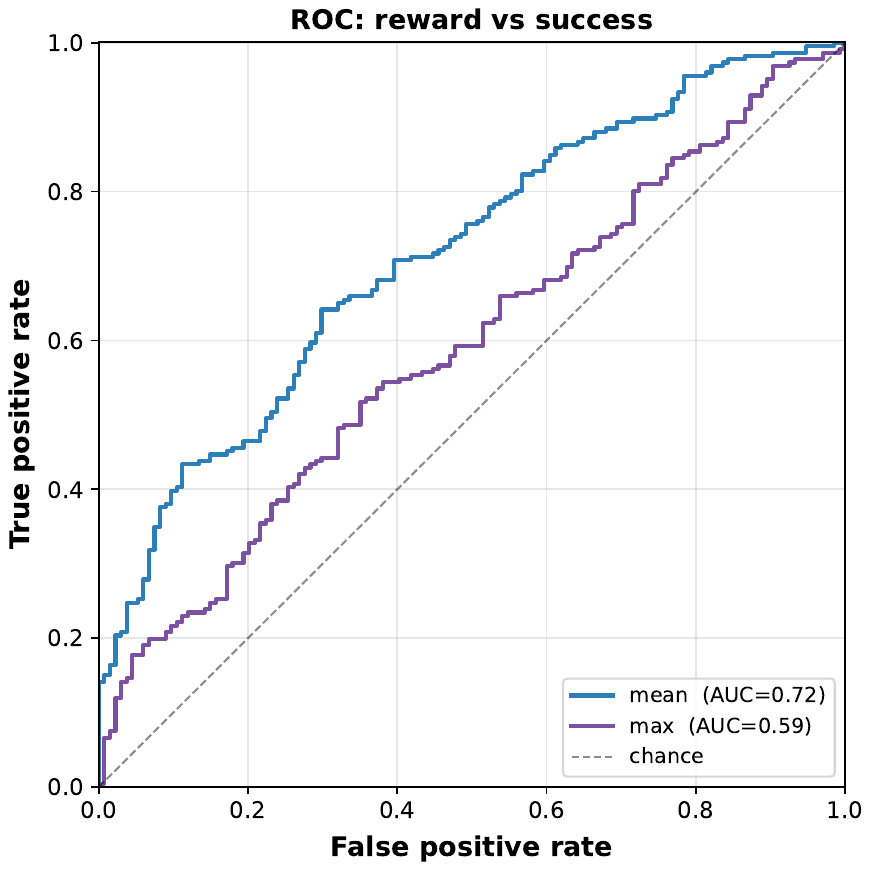}
    \caption{ROC for predicting success from the critic reward. Mean
    aggregation (AUC $=0.72$) clearly outperforms single-frame max
    (AUC $=0.59$).}
    \label{fig:vlm-critic-auroc}
\end{subfigure}
\caption{Critic reward predicts task success. Calibration by reward quartile
(left) and ROC comparing reward aggregations (right) over 360 rollouts.}
\label{fig:critic}
\end{figure}


\noindent
\textbf{Effect of different VLM critic choices. }
To determine the most effective reward signal for guiding our diffusion video generator, we ablate two distinct VLM critic architectures: CycleReward \cite{bahng2025cycle} and SBERT-based embeddings. The critic is responsible for providing high-level reasoning feedback by measuring the semantic alignment between the generated video frames and the goal state.

As illustrated in Figure \ref{fig:vlm-critic}, we observe that the optimal critic choice varies by task geometry and semantic complexity. For instance, CycleReward shows a slight edge in highly structured tasks like Coffee SetupMug, whereas SBERT significantly outperforms in varied PnP scenarios such as PnP CounterToSink and PnP MicrowaveToCounter. Ultimately, we select SBERT as our primary reasoning critic because it yields a higher average success rate across the full task suite, indicating a more stable semantic representation for evaluating environmental transitions.



\begin{figure}[t]
\centering

\begin{subfigure}{0.7\linewidth}
    \centering
    \includegraphics[width=\linewidth]{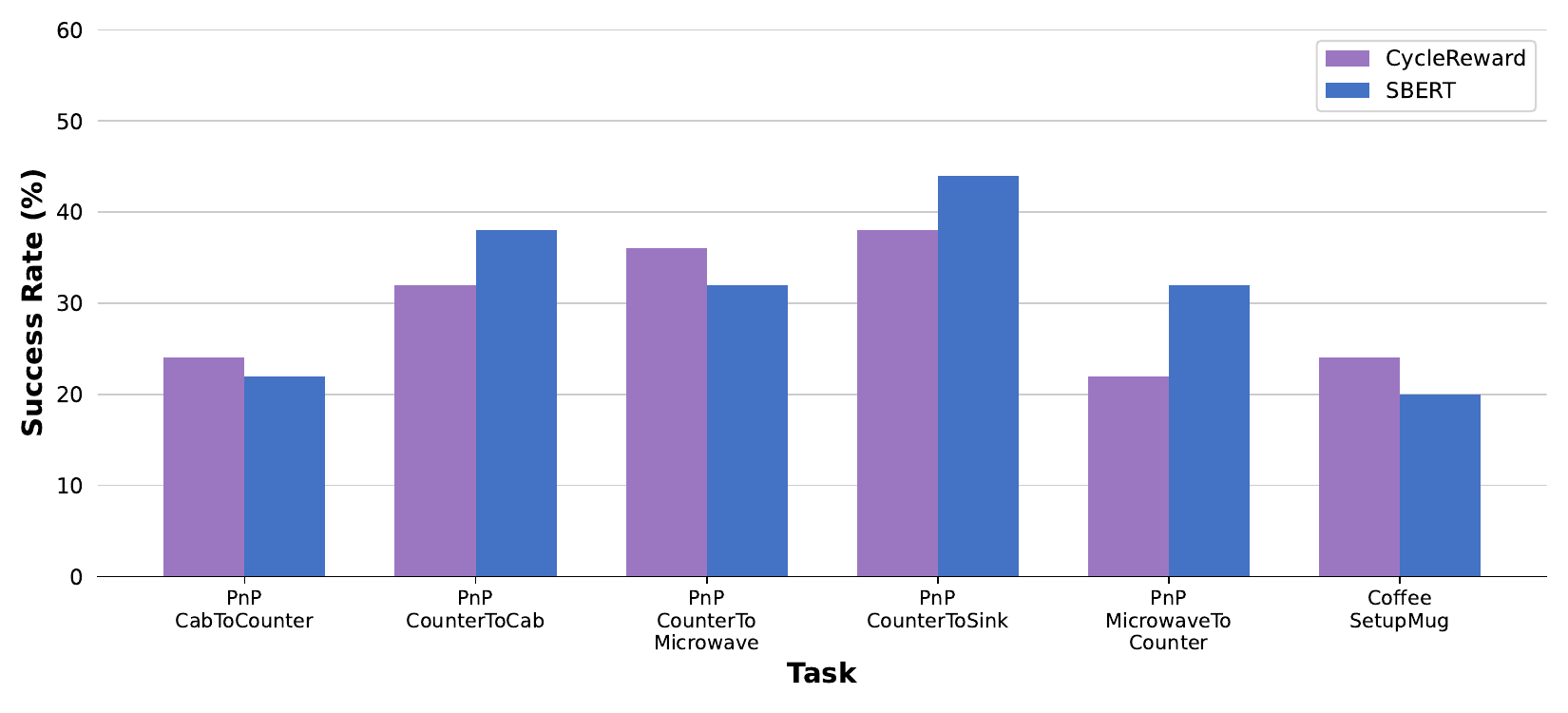}
    \caption{Success rate across individual tasks.}
    \label{fig:vlm-critic-tasks}
\end{subfigure}
\hfill
\begin{subfigure}{0.25\linewidth}
    \centering
    \includegraphics[width=\linewidth]{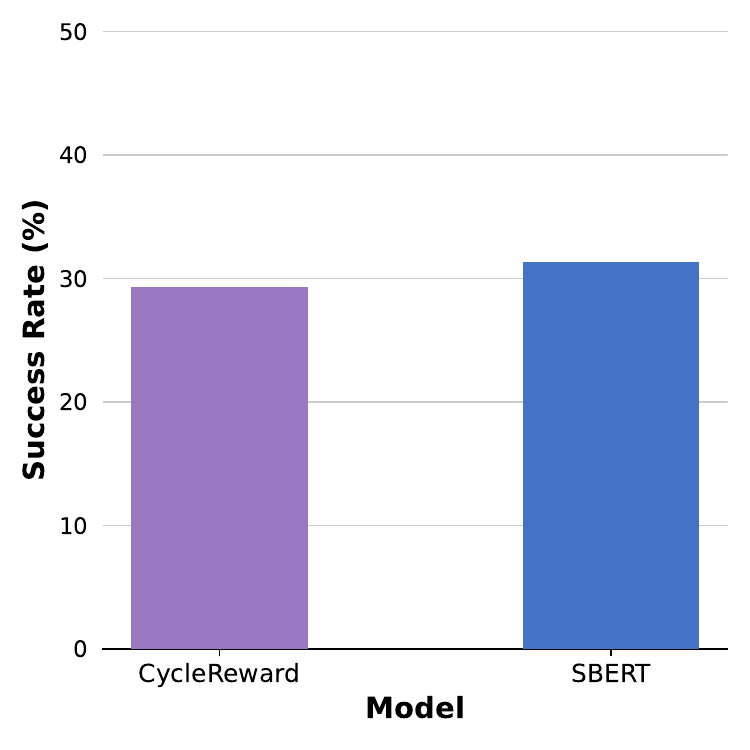}
    \caption{Average success rate across 6 tasks.}
    \label{fig:vlm-critic-avg}
\end{subfigure}

\caption{Performance comparison of different VLM-based reward models for reasoning guidance. We evaluate the impact of CycleReward~\cite{bahng2025cycle} versus BLIP-SBERT score \cite{black2023ddpo} on success rates across six challenging manipulation tasks including five Pick-and-Place (PnP) variants and a mug setup task as shown in (a). While performance is task-dependent, SBERT variant demonstrates slightly superior robustness and a higher overall average success rate as shown in (b).}
\label{fig:vlm-critic}
\end{figure}
\begin{figure}[!t]
  \centering
    \includegraphics[width=0.8\textwidth]{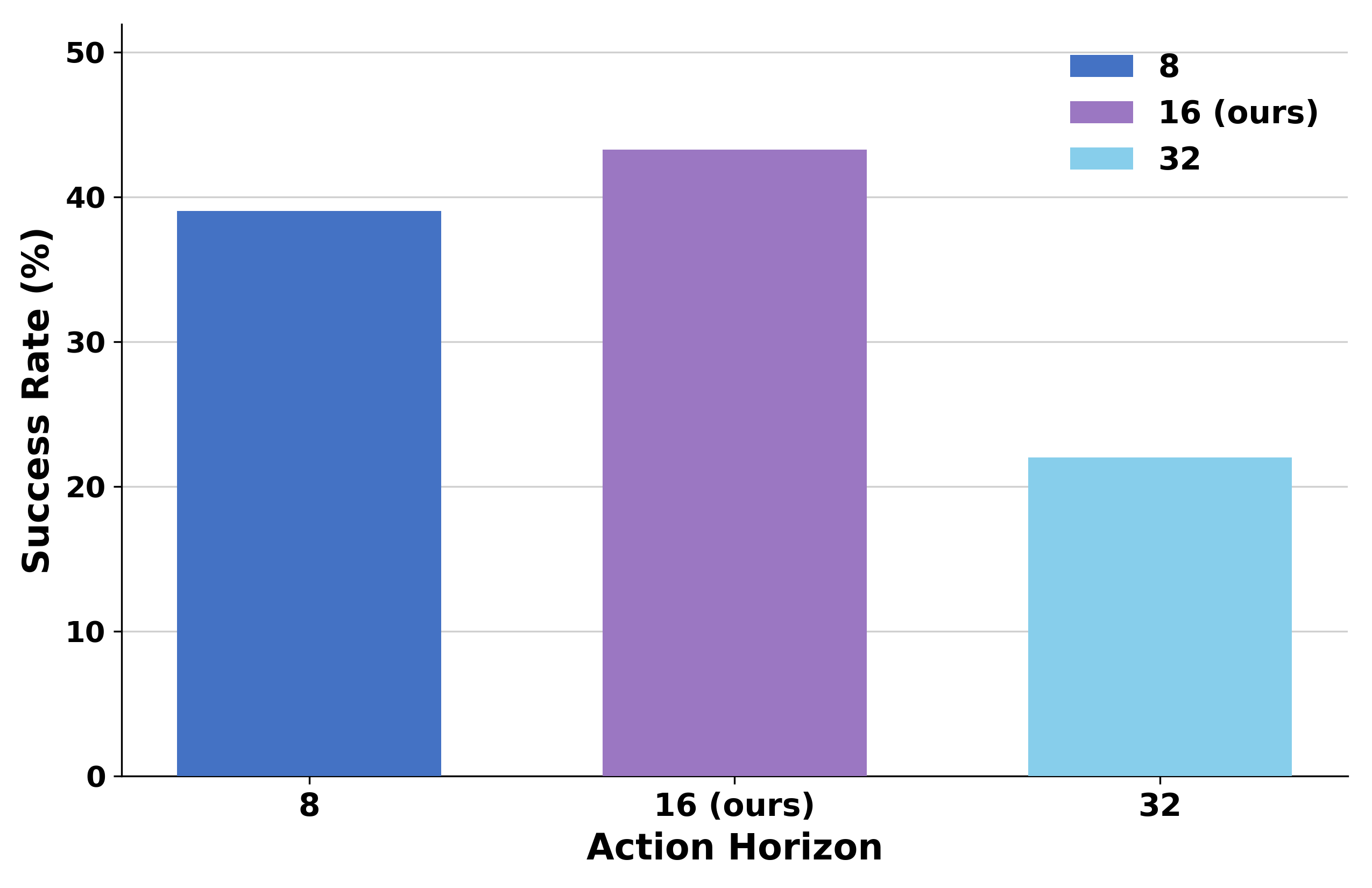}
\caption{\textbf{Effect of changing action horizon. }Our choice of action horizon 16 performs best overall computed across 3 tasks with 50 demos each.}
    \label{fig:action-horizon}
\end{figure}

\medskip
\noindent
\textbf{Effect of changing action horizon. }We perform evaluation of our model changing the action horizon on 150 demos across 3 tasks. As shown in Fig. \ref{fig:action-horizon}, we observe a clear non-monotonic effect: horizon 16 (ours) performs best while a shorter horizon 8 drops the performance and a longer horizon 32 drops the performance further. This suggests horizon 16 gives the best tradeoff between re-planning frequency and action commitment; very short horizons can make control too reactive, while very long horizons increase open-loop drift and error accumulation.


\medskip
\noindent
\textbf{Joint Training vs. Stage-wise Training. }As described in the main paper, RoboTALES adopts a coupled, single-stage joint training strategy in which the video generator and action policy are optimized simultaneously. This end-to-end design allows action-level gradients to flow back into the video generator's decoder layers, encouraging it to produce latent representations that are not only visually faithful and semantically aligned but also directly informative for downstream control. We verify this design choice by comparing the decoupled stage-wise training with our proposed joint training in Fig. \ref{fig:joint-stagewise}. As is obvious from the plot, our proposed joint training pipeline promotes tighter co-adaptation between imagination and action, leading to overall improvements in  action prediction.
\begin{figure}[t]
  \centering
    \includegraphics[width=1.0\textwidth]{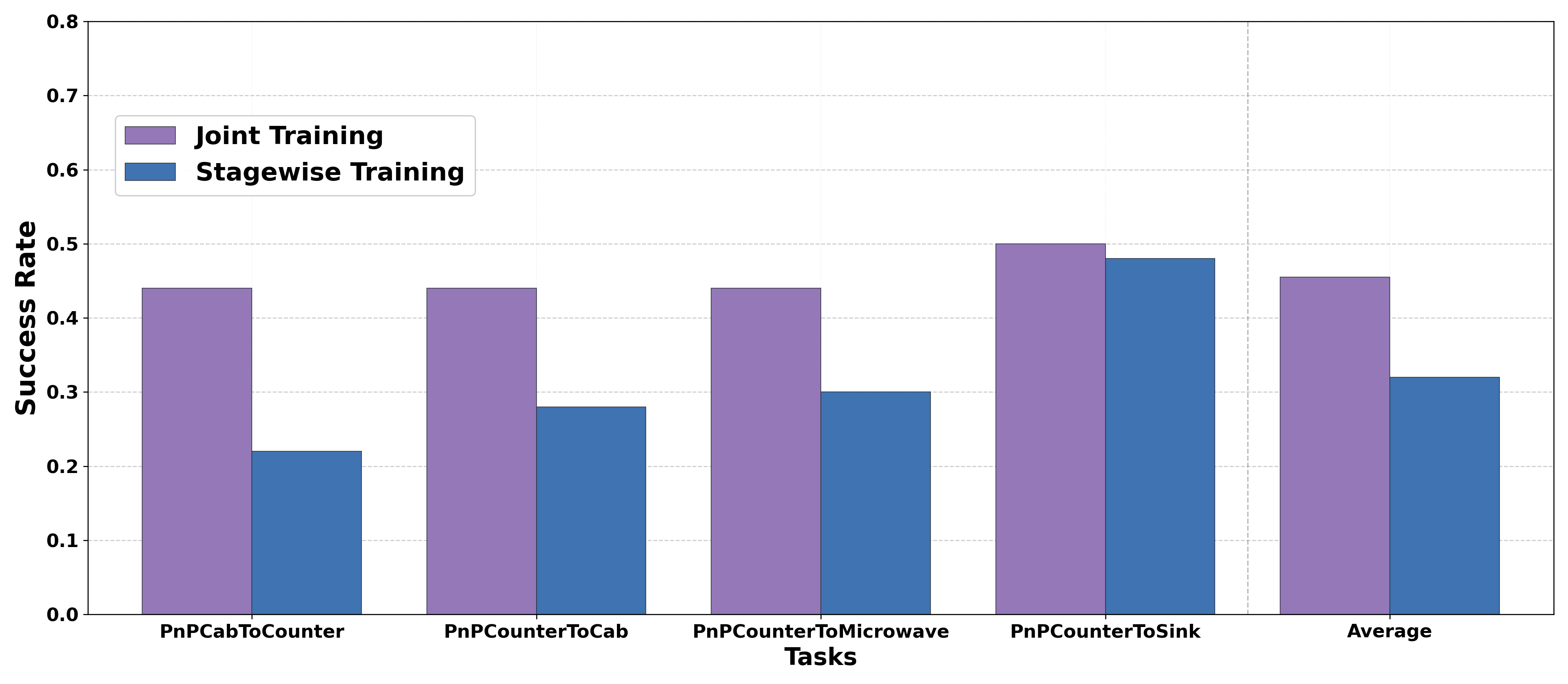}
\caption{Joint training vs Stage-wise training: Our proposed joint training setup shows enhanced task-wise success-rates compared to stage-wise training computed across 4 RoboCasa tasks. The Average scores (right) further strengthen our claim.}
    \label{fig:joint-stagewise}
\end{figure}

\noindent
\textbf{Planner Decomposition Quality. }
To assess planning quality, we perform a blind pairwise evaluations on 50 randomly sampled task instructions using GPT-4o as a judge, comparing our planner-generated decompositions against a prompt-only baseline in Fig. \ref{fig:planner-decomposition}. For each instruction, the decomposition is judged on following 5 criteria: correctness, completeness, ordering, executability, and non-redundancy. The left plot (pairwise winner rate) shows a clear preference for the planner output: it is selected as better in 60\% of comparisons, versus 38\% for prompt-only, with only 2\% ties. The right plot (decomposition length) explains an important part of this gain: planner outputs are substantially shorter on average (2.80 steps vs 5.28), indicating less redundant and more compact task structure. Taken together, the two plots suggest that the planner is not just producing longer, more verbose plans to “game” the judge; instead, it generates concise decompositions that are judged better overall for task execution quality. This supports the claim that planner structure improves decomposition usefulness and efficiency relative to direct prompt-only generation.

\begin{figure}[H]
  \centering
    \includegraphics[width=1.0\textwidth]{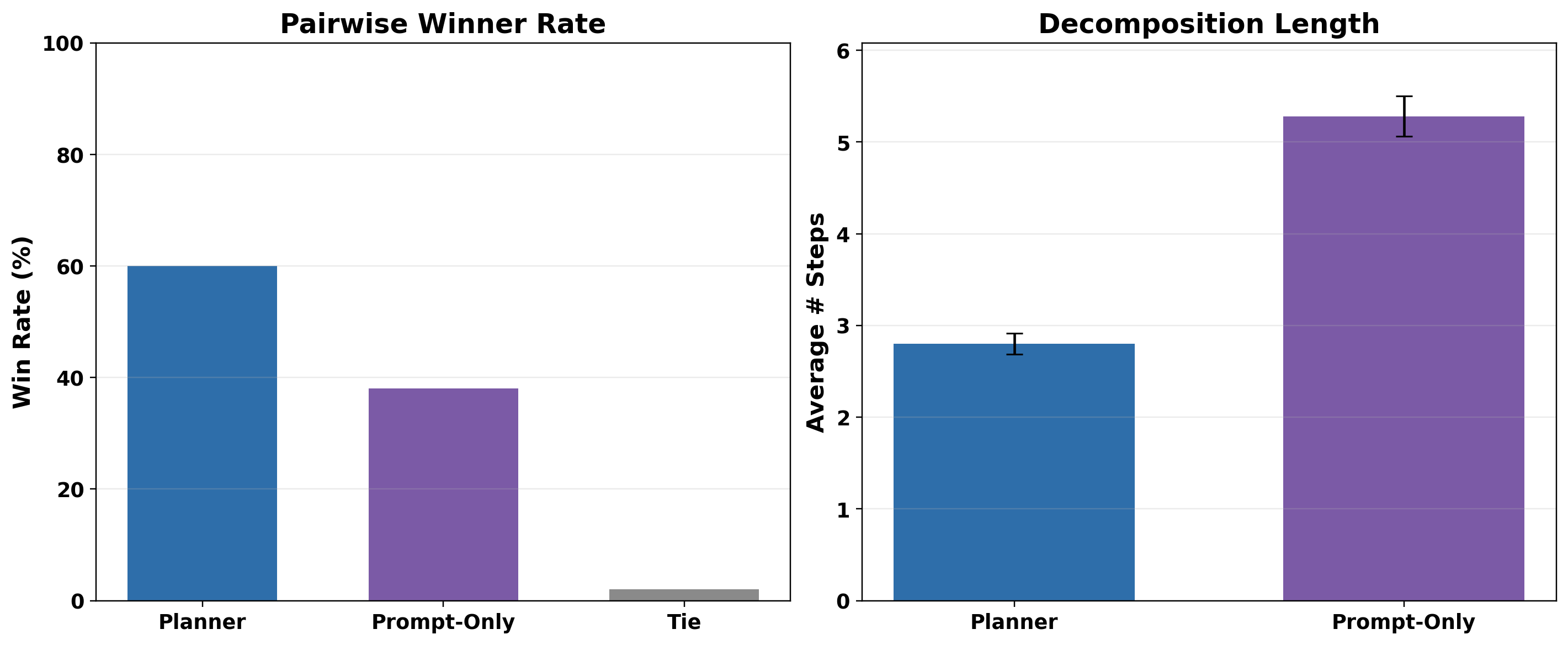}
\caption{LLM-judge evaluation of decomposition quality (N=50): planner decompositions outperform prompt-only (60\% vs 38\% wins, 2\% ties) while using fewer steps (2.80 vs 5.28), indicating better plan efficiency and overall preference.}
    \label{fig:planner-decomposition}
\end{figure}

\noindent
\textbf{Retaining and Improving General Video Generation Capabilities. }
To validate whether our joint task-aligned training preserves the underlying generative video prior, we evaluate the original SVD model and our jointly trained video model on non-robotics real videos (from UCF101 dataset), using 200 matched conditioning samples and 30 denoising steps. We compare generated futures to ground-truth future clips in feature space. As shown in Fig.~\ref{fig:video-gen}, our model is consistently closer to real-video dynamics: Fréchet distance decreases from 0.6715 to 0.2437 (63.7\% lower), MMD-RBF decreases from 0.2419 to 0.0631 (73.9\% lower), and per-sample cosine alignment to ground truth increases from 0.5955 to 0.8535 (+0.258 absolute, 43.3\% relative). Distributional fidelity also improves: diversity-ratio-to-GT moves from 0.7201 (under-dispersed) to 1.0249 (near ideal), and covariance-trace ratio improves from 0.5418 to 1.0494, indicating much less variance collapse after control training. Overall, these results show that joint training does not collapse SVD into a narrow task-specific predictor; instead, it retains and improves general real-world video generation quality relative to the original SVD baseline.

\begin{figure}[t]
  \centering
    \includegraphics[width=1.0\textwidth]{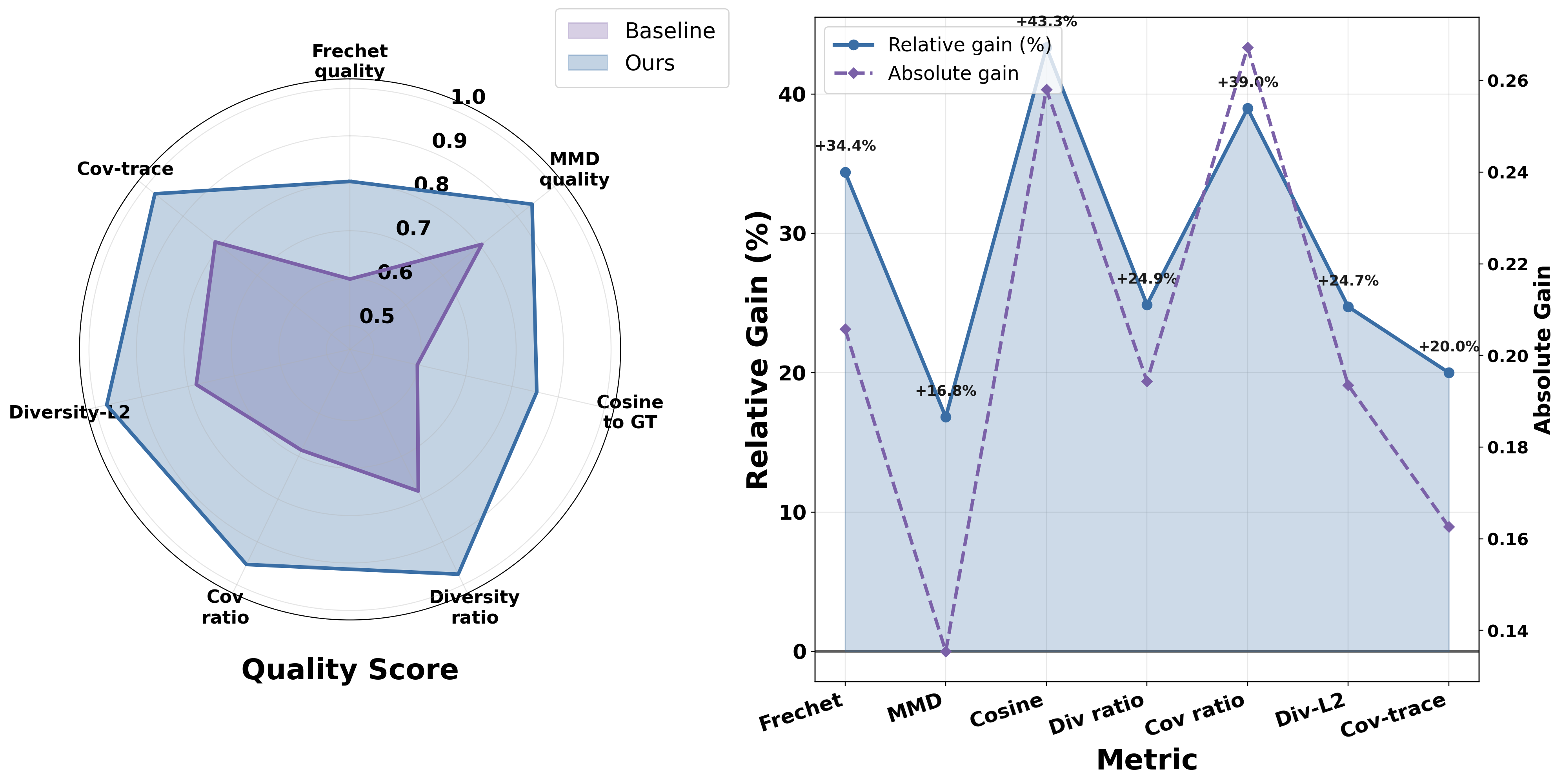}
\caption{\textbf{Generative retention on UCF101.} We compare the original SVD with our jointly trained video model using 200 matched conditioning samples from UCF101 datset with 30 denoising steps each, evaluated against ground-truth future clips. The radar plot shows absolute quality profiles across alignment and distributional metrics, while the gain-area plot reports per-metric improvements (relative gain \% with absolute-gain overlay). Our model improves all metrics, with the largest gains in Fréchet and MMD alignment, and substantially better diversity/covariance matching to the real-video distribution.}

    \label{fig:video-gen}
\end{figure}

\subsection{Additional Qualitative Results. }
\label{appendix:additional-qualitative}
We present extended qualitative comparisons and visualizations to further demonstrate the robustness of our framework. Figure~\ref{fig:visual-result-1} highlights our method's ability to maintain goal-directed behavior over long horizons compared to the baseline, while Figure~\ref{fig:visual-result-2} showcases successful execution across a diverse array of complex kitchen manipulation tasks.

\begin{figure}[t]
  \centering
    \includegraphics[width=1.0\textwidth]{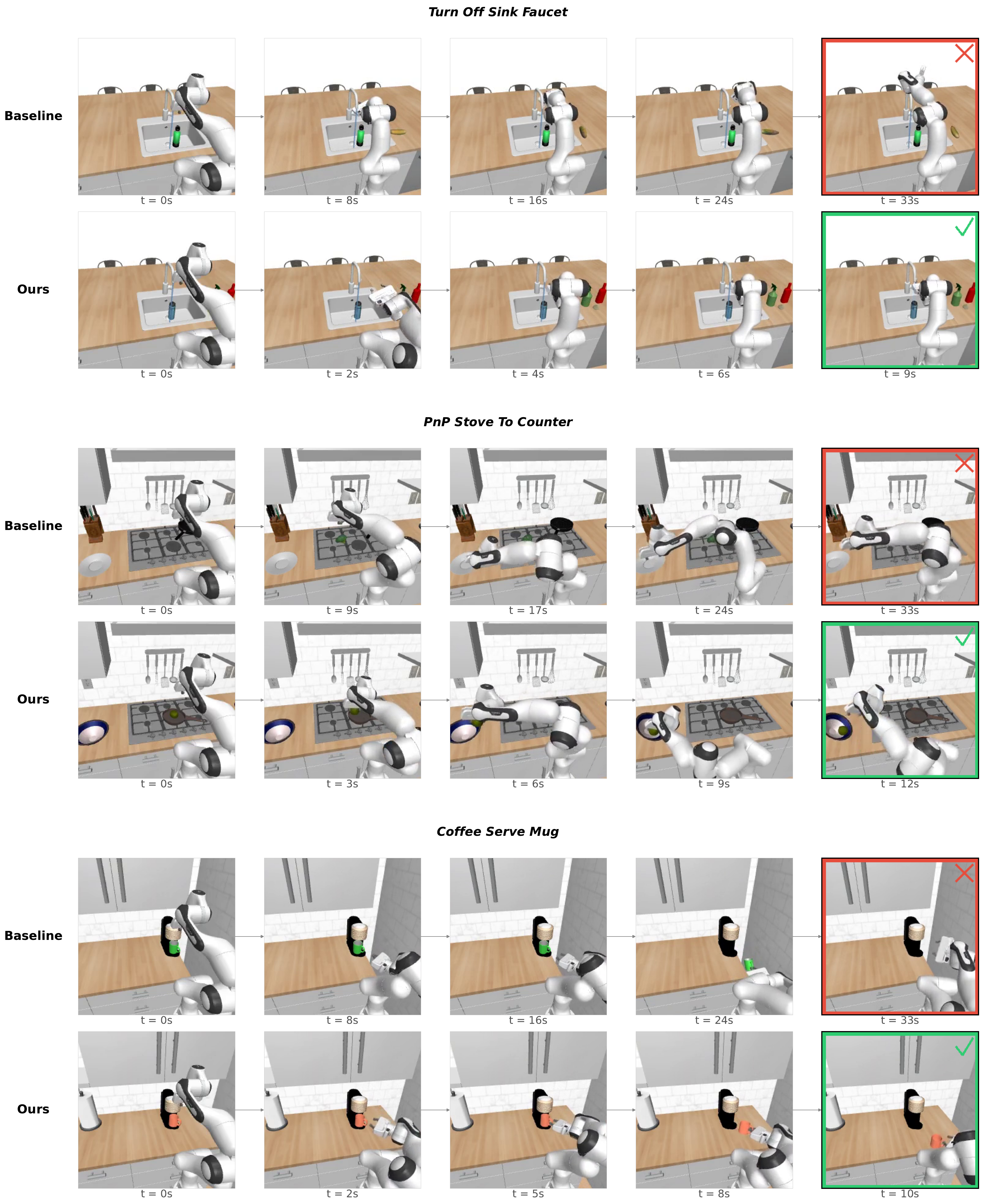}
    \caption{Comparing our method to the baseline across different tasks . We show how our model performs against the baseline on tasks like turning off a faucet, moving items from a stove to a counter, and serving coffee. In each case, the baseline (top rows) loses track of the goal over time, leading to distorted movements or failed actions. Our method (bottom rows) uses high-level planning to stay on track, completing the tasks accurately and more quickly.}
    \label{fig:visual-result-1}
\end{figure}

\begin{figure}[t]
  \centering
    \includegraphics[width=1.0\textwidth]{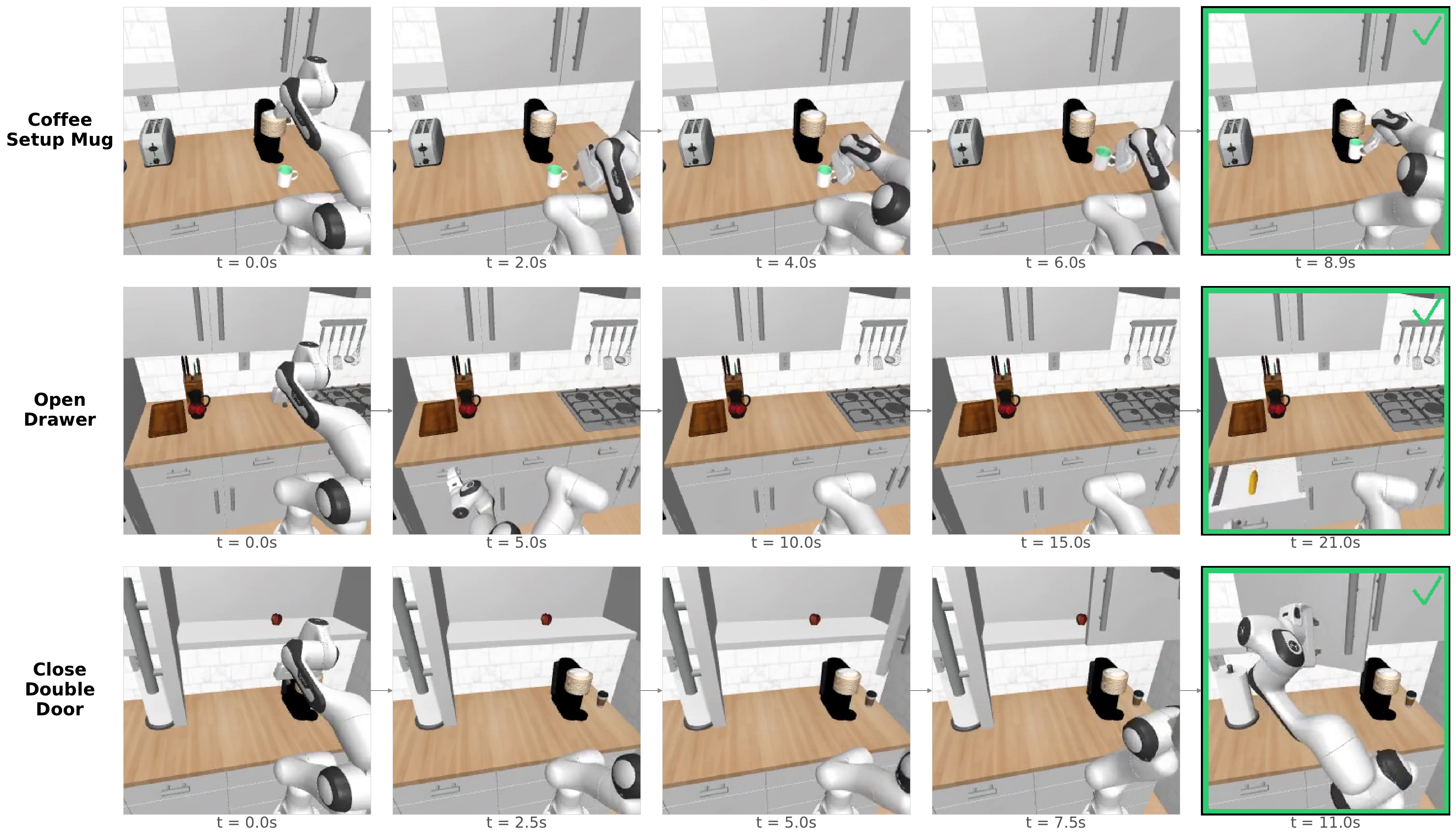}
    \caption{Success of our method in various kitchen scenarios . This figure demonstrates our model’s ability to handle different complex tasks, such as setting up a coffee mug, opening drawers, and closing double doors. Our model consistently creates smooth, goal-oriented movements while managing multiple camera views and complex objects. These successful results show that combining smart reasoning with a video generator leads to more reliable robot performance in real-world settings}
    \label{fig:visual-result-2}
\end{figure}

\subsection{Details about Planner}
\label{appendix:planner_details}

To convert high-level RoboCasa task instructions into executable sequences for video rollout, we utilize \texttt{Gemini 2.5 Pro} as a offline task planner. The planner follows the \textit{VLWM-style} abstraction, ensuring each step represents a meaningful and observable transition in the environment state. We impose a constraint on the planner to generate between 2 and 5 action steps per instruction. This range ensures that atomic tasks are sufficiently decomposed for the policy to learn meaningful sub-goals, while complex composite tasks are compressed into high-level transitions that remain manageable for the video rollout horizon.

To ensure computational efficiency during training and consistency across inference rollouts, we pre-generate these decompositions for the entire dataset. These steps are stored in a persistent local cache, indexed by a unique hash of the instruction and model version. This avoids redundant LLM API calls and ensures that the downstream policy is always trained on a stable set of sub-goals.

Below, we provide the system prompt employed to obtain the decompositions of the task instruction. 

\begin{tcolorbox}[promptbox, title={\faRobot\quad System Prompt}]
You are a planner that converts ONE RoboCasa task instruction
into a short, ordered list of ACTIONS for video rollout.

\medskip
\medskip

Planning principle

- Follow VLWM-style abstraction: each step is an ACTION that causes a meaningful, observable world-state change. Keep state implicit.

\medskip
\medskip

Hard constraints

- Always return between 2 and 5 actions inclusive.
- Output ONLY a strict JSON object with this shape:
  {"steps": ["<action 1>", "<action 2>", "..."]}
- Each action is a concise, imperative verb phrase (<= 12 words).
- Preserve explicit ordering from the instruction.

\medskip
\medskip

Granularity rules

- Composite instructions: 3-5 actions capturing key transitions.

- Atomic instructions: Split into two micro-actions (e.g., reach/align).
\end{tcolorbox}

\vspace{6pt}

\begin{tcolorbox}[promptbox, 
  colframe=black!50, 
  colback=blue!3,
  boxed title style={colback=black!55},
  title={\faListOl\quad Instruction Prompt \& In-Context Examples}]
Task instruction:

\{\{SYSTEM\_PROMPT\}\}

\medskip
\medskip
Return ONLY a strict JSON object of 2-5 concise action strings:
{"steps": ["...", "..."]}

\medskip
\medskip
In-context examples:

Instruction:
  Open the microwave, put the bowl inside, close it, then start it.

Expected:

  {"steps": [
  
    "open the microwave door",
    
    "place the bowl inside the microwave",
    
    "close the microwave door",
    
    "press the start button"
    
  ]}
\end{tcolorbox}

Table~\ref{tab:task_decomposition_examples} showcases how the \texttt{Gemini 2.5 Pro} planner translates high-level semantic instructions into sequential, observable sub-goals while adhering to a 2--5 step granularity constraint. This cached decomposition enables consistent, multi-stage task execution by grounding complex human requests into a manageable sequence of intermediate world-state transitions.

\begin{table}[H]
\centering
\small
\renewcommand{\arraystretch}{1.2}
\caption{Examples of task instructions and their corresponding planner-generated step decompositions for the Libero10 dataset}
\begin{tabularx}{\linewidth}{p{5.2cm} X}
\toprule
\textbf{Original Task Instruction} & \textbf{Decomposed Steps} \\
\midrule

Turn on the stove and put the moka pot on it &
1. Turn on the stove \newline
2. Place the moka pot on the stove burner \\
\hline
Put the black bowl in the bottom drawer of the cabinet and close it &
1. Pick up the black bowl \newline
2. Open the cabinet bottom drawer \newline
3. Place the black bowl in the drawer \newline
4. Close the cabinet bottom drawer \\
\hline
Put the yellow and white mug in the microwave and close it &
1. Put the yellow and white mug in the microwave \newline
2. Close the microwave door \\
\hline
Put both moka pots on the stove &
1. Place the first moka pot on the stove \newline
2. Place the second moka pot on the stove \\
\hline
Put both the alphabet soup and the cream cheese box in the basket &
1. Pick up the alphabet soup \newline
2. Place the alphabet soup in the basket \newline
3. Pick up the cream cheese box \newline
4. Place the cream cheese box in the basket \\
\hline
Put both the alphabet soup and the tomato sauce in the basket &
1. Place the alphabet soup in the basket \newline
2. Place the tomato sauce in the basket \\
\hline
Put both the cream cheese box and the butter in the basket &
1. Place the cream cheese box in the basket \newline
2. Place the butter in the basket \\
\hline
Put the white mug on the left plate and put the yellow and white mug on the right plate &
1. Put the white mug on the left plate \newline
2. Put the yellow and white mug on the right plate \\
\hline
Put the white mug on the plate and put the chocolate pudding to the right of the plate &
1. Put the white mug on the plate \newline
2. Place the chocolate pudding to the right of the plate \\
\hline
Pick up the book and place it in the back compartment of the caddy &
1. Pick up the book \newline
2. Place the book in the caddy back compartment \\
\bottomrule
\end{tabularx}
\label{tab:task_decomposition_examples}

\end{table}


\subsection{Limitations}
\label{appendix:limitations}
While RoboTALES demonstrates strong performance across diverse manipulation tasks, it is worth noting certain limitations. First, our VLM Critic is frozen throughout training and provides a relatively coarse, task-level reward signal. A learnable reward model that co-adapts with the video generator and action policy could provide richer, more nuanced feedback -- particularly for contact-rich tasks where subtle differences in grasp quality or object pose are difficult to assess from pixel-level text-image similarity alone.
Second, our framework inherits the computational cost of video diffusion models. The DDPO-based policy optimization requires multiple denoising rollouts per training step to estimate gradients, making training expensive. Third, our setup purely simulation-based in current state. We are actively working on addressing these limitations as part of our ongoing and future work.

\end{document}